\PassOptionsToPackage{table}{xcolor}
\documentclass{article} % For LaTeX2e
\usepackage{iclr2022_conference,times}

\usepackage{amsmath,amsfonts,bm}

\def\eqref#1{equation~\ref{#1}}
\def\1{\bm{1}}

\DeclareMathAlphabet{\mathsfit}{\encodingdefault}{\sfdefault}{m}{sl}
\SetMathAlphabet{\mathsfit}{bold}{\encodingdefault}{\sfdefault}{bx}{n}

\usepackage{hyperref}
\usepackage{url}

\usepackage{times}
\usepackage{epsfig}
\usepackage{graphicx}
\usepackage{amsmath}
\usepackage{amssymb}
\usepackage{multirow}
\usepackage{caption}
\usepackage{makecell}
\usepackage{subcaption}
\usepackage{floatrow}
\usepackage{algorithm}
\usepackage[noend]{algpseudocode}
\usepackage[accsupp]{axessibility}
\graphicspath{ {images/} {images-suppl/} }

\iclrfinalcopy

\newcommand\norm[1]{\left\lVert#1\right\rVert}
\newfloatcommand{capbtabbox}{table}[][\FBwidth]

\definecolor{Gray}{gray}{0.93}
\newcolumntype{g}{>{\columncolor{Gray}}c}
\definecolor{Gray2}{gray}{0.85}
\newcolumntype{h}{>{\columncolor{Gray2}}c}
\definecolor{Gray3}{gray}{0.8}
\newcolumntype{a}{>{\columncolor{Gray3}}c}
\definecolor{Gray4}{gray}{0.97}
\newcolumntype{e}{>{\columncolor{Gray4}}c}

\title{Divergence-aware Federated Self-Supervised Learning}

\author{Weiming Zhuang$^{1,3}$, \, Yonggang Wen$^{2}$, \, Shuai Zhang$^{3}$ \\
$^{1}$S-Lab, NTU, Singapore\, $^{2}$NTU, Singapore\, $^{3}$SenseTime Research \\
\texttt{weiming001@e.ntu.edu.sg,ygwen@ntu.edu.sg,zhangshuai@sensetime.com} \\
}

\begin{document}

\maketitle

\begin{abstract}

Self-supervised learning (SSL) is capable of learning remarkable representations from centrally available data. Recent works further implement federated learning with SSL to learn from rapidly growing decentralized unlabeled images (e.g., from cameras and phones), often resulted from privacy constraints. Extensive attention has been paid to SSL approaches based on Siamese networks. However, such an effort has not yet revealed deep insights into various fundamental building blocks for the federated self-supervised learning (FedSSL) architecture. We aim to fill in this gap via in-depth empirical study and propose a new method to tackle the non-independently and identically distributed (non-IID) data problem of decentralized data. Firstly, we introduce a generalized FedSSL framework that embraces existing SSL methods based on Siamese networks and presents flexibility catering to future methods. In this framework, a server coordinates multiple clients to conduct SSL training and periodically updates local models of clients with the aggregated global model. Using the framework, our study uncovers unique insights of FedSSL: 1) stop-gradient operation, previously reported to be essential, is not always necessary in FedSSL; 2) retaining local knowledge of clients in FedSSL is particularly beneficial for non-IID data. Inspired by the insights, we then propose a new approach for model update, Federated Divergence-aware Exponential Moving Average update (FedEMA). FedEMA updates local models of clients adaptively using EMA of the global model, where the decay rate is dynamically measured by model divergence. Extensive experiments demonstrate that FedEMA outperforms existing methods by 3-4\% on linear evaluation. We hope that this work will provide useful insights for future research.

\end{abstract}

% These methods have empowered large companies to improve internal systems \citep{yann2021facebook}, heavily relying on the assumption that images are centrally available in cloud servers, such as public data on the Internet.

% researchers have designed many proxy tasks for the networks to learn from the images by completing these tasks, such as rotation, image inpainting. The state-of-the-art methods employ contrastive learning that minimizes the similarity of two augmented views of the same image and pushing apart two different images. These methods have empowered large companies such as Facebook to improve their content understanding systems using the data that are centrally available, such as crawled from the Internet, or Instagrams. 

\section{Introduction}

Self-supervised learning (SSL) has attracted extensive research interest for learning representations without relying on expensive data labels. In computer vision, the common practice is to design proxy tasks to facilitate visual representation learning from unlabeled images \citep{doersch2015unsup-context-prediction, noroozi2016jigsaw, zhang2016colorful-image,gidaris2018unsup-image-rotation}. Among them, the state-of-the-art SSL methods employ contrastive learning that uses Siamese networks to minimize the similarity of two augmented views of images \citep{wu2018unsupervised-instance, chen2020simclr, he2020moco, grill2020byol, chen2020simsiam}. All these methods heavily rely on the assumption that images are centrally available in cloud servers, such as public data on the Internet.

However, the rapidly growing amount of decentralized images may not be centralized due to increasingly stringent privacy protection regulations \citep{gdpr}. The increasing number of edge devices, such as street cameras and phones, are generating a large number of unlabeled images, but these images may not be centralized as they could contain sensitive personal information like human faces. Besides, learning representations from these images could be more beneficial for downstream tasks deployed in the same scenarios \citep{yan2020neural}. A straightforward method is to adopt SSL methods for each edge, but it results in poor performance \citep{zhuang2021fedu} as decentralized data are mostly non-independently and identically distributed (non-IID) \citep{Li2020FedChallenges}.

Federated learning (FL) has emerged as a popular privacy-preserving method to train models from decentralized data \citep{fedavg}, where clients send training updates to the server instead of raw data. The majority of FL methods, however, are not applicable for unsupervised representation learning because they require fully labeled data \citep{caldas2018leaf}, or partially labeled data in either the server or clients \citep{jin2020fed-unlabeled-survey,jeong2021fedsemi}. Recent studies implement FL with SSL methods that are based on Siamese networks, but they only focus on a single SSL method. For example, FedCA \citep{zhang2020fedca} is based on SimCLR \citep{chen2020simclr} and FedU \citep{zhuang2021fedu} is based on BYOL \citep{grill2020byol}. These efforts have not yet revealed deep insights into the fundamental building blocks of Siamese networks for federated self-supervised learning.

In this paper, we investigate the effects of fundamental components of federated self-supervised learning (FedSSL) via in-depth empirical study. To facilitate fair comparison, we first introduce a generalized FedSSL framework to embrace existing SSL methods that differ in building blocks of Siamese networks. The framework comprises of a server and multiple clients: clients conduct SSL training using Siamese networks --- an online network and a target network; the server aggregates the trained online networks to obtain a new global network and uses this global network to update the online networks of clients in the next round of training. FedSSL primarily focuses on the cross-silo FL where clients are stateful with high availability \citep{kairouz2019advances}. 

We conduct empirical studies based on the FedSSL framework and discover important insights of FedSSL. Among four popular SSL methods (SimCLR \citep{chen2020simclr}, MoCo \citep{he2020moco}, BYOL \citep{grill2020byol}, and SimSiam \citep{chen2020simsiam}, FedBYOL achieves the best performance, whereas FedSimSiam yields the worst performance. More detailed analysis uncover the following unique insights: 1) Stop-gradient operation, essential for SimSiam and BYOL, is not always essential in FedSSL; 2) Target networks of clients are essential to gain knowledge from online networks; 3) Keeping local knowledge of clients is beneficial for performance on non-IID data.

Inspired by the insights, we propose a new approach, Federated Divergence-aware Exponential Moving Average update (FedEMA) \footnote{Intuitively, FedSSL is analogous to a SuperClass in object-oriented programming (OOP), then FedEMA is a SubClass that inherits FedSSL and overrides the model update method.} , to address the non-IID data problem. Specifically, instead of updating online networks of clients simply by the global network, FedEMA updates them via exponential moving average (EMA) of the global network, where the decay rate of EMA is measured by the divergence of global and online networks dynamically. Extensive experiments demonstrate that FedEMA outperforms existing methods in a wide range of settings. We believe that important insights from this study will shed light on future research. Our main contributions are threefold:

\begin{itemize}
   \item We introduce a new generalized FedSSL framework that embraces existing SSL methods based on Siamese networks and presents flexibility catering to future methods.
   \item We conduct in-depth empirical studies of FedSSL based on the framework and discover deep insights of the fundamental building blocks of Siamese networks for FedSSL.
   %  that are benchmark the existing SSL methods in federated settings and gain important insights on the impact factors for FedSSL.
   \item Inspired by the insights, we further propose a new model update approach, FedEMA, that adaptively updates online networks of clients with EMA of the global network. Extensive experiments show that FedEMA outperforms existing methods in a wide range of settings.
   % \item We validate the D-EMA with extensive experiments on a wide range of settings, especially focusing on the non-IID settings. 
\end{itemize}

\section{Related Work}
\label{sec:related-work}

% Siamese networks.
% contrastive learning: contrastive and non-contrastive methods. 

\textbf{Self-supervised Learning} \, In computer vision, self-supervised learning (SSL) aims to learn visual representations without any labels. Discriminative SSL methods facilitate learning with proxy tasks \citep{pathak2016inpainting,noroozi2016jigsaw,zhang2016colorful-image,gidaris2018unsup-image-rotation}. Among them, contrastive learning \citep{oord2018contrastive-predictive-coding, bachman2019contrastive} has become a promising principle. It uses Siamese networks to minimize the similarity of two augmented views (positive pairs) and maximize the similarity of two different images (negative pairs). These methods are either contrastive or non-contrastive ones: \textit{contrastive} SSL methods require negative pairs \citep{chen2020simclr, he2020moco} to prevent training collapse; \textit{non-contrastive} SSL methods \citep{grill2020byol,chen2020simsiam} are generally more efficient as they maintain remarkable performances using only positive pairs. However, these methods do not perform well on decentralized non-IID data \citep{zhuang2021fedu}. We analyze their similarities and variances and propose a generalized FedSSL framework.
% \textit{contrastive} SSL methods require negative pairs, from a large batch size \citep{chen2020simclr} or a memory bank \citep{he2020moco}, to prevent training collapse;

\textbf{Federated Learning} \, Federated learning (FL) is a distributed training technique for learning from decentralized parties without transmitting raw data to a central server \citep{fedavg}. Among many studies that address the non-IID data challenge \citep{zhao2018non-iid,fedprox,Wang2020fedma,zhuang2021fedureid}, Personalized FL (PFL) aims to learn personalized models for clients \citep{tan2021personlized-fl-survey}. Although some PFL methods interpolate global and local models \citep{hanzely2020lower,mansour2020three,deng2021adaptive}, our proposed FedEMA differ in the motivation, application scenario, and measurement of the decay rate. Besides, the majority of existing works only consider supervised learning where clients have fully labeled data. Although recent works propose federated semi-supervised learning \citep{jin2020survey-fedsemi,zhang2020benchmark-fedsemi,jeong2021fedsemi} or federated domain adaptation \citep{peng2020fuda,zhuang2021fedfr}, they still need labels in either the server or clients. This paper focuses on purely unlabeled decentralized data.

\textbf{Federated Unsupervised Learning} \, Learning representations from unlabeled decentralized data while preserving data privacy is still a nascent field. Federated unsupervised representation learning is first proposed by \citet{van2020fedae} based on autoencoder, but it neglects the non-IID data challenge. \citet{zhang2020fedca} address the non-IID issue with potential privacy risk for sharing features. Although \citet{zhuang2020fedreid} address the issue based on BYOL as our FedEMA, they do not shed light on why BYOL works best. Since SSL methods are evolving rapidly and new methods are emerging, we introduce a generalized FedSSL framework and deeply investigate the fundamental components to build up practical guidelines for the generic FedSSL framework.

\begin{figure*}[t]
   \begin{center}
   \includegraphics[width=1\linewidth]{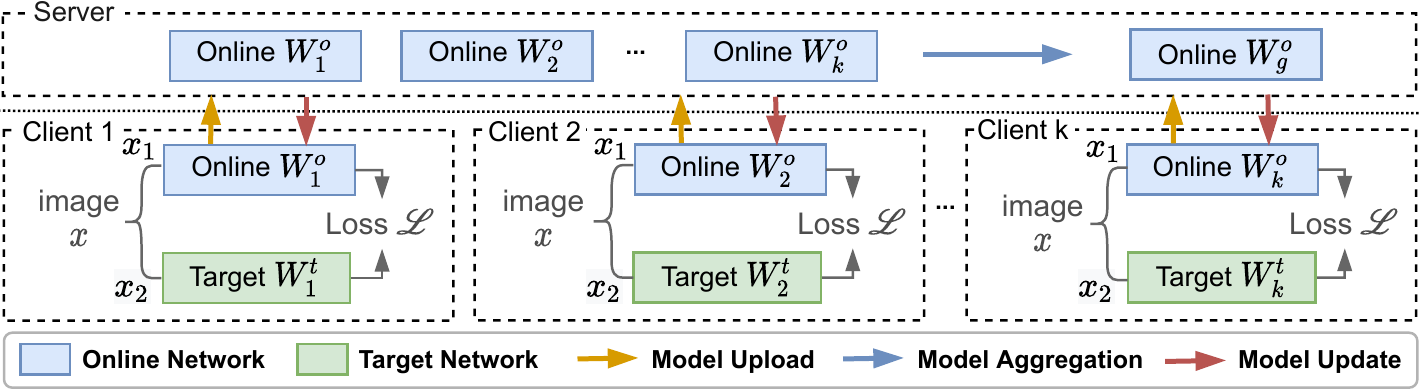}
      \caption{Overview of federated self-supervised learning (FedSSL) framework. It comprises an end-to-end training pipeline with four steps: 1) Each client $k$ conducts local training on unlabeled data $\mathcal{D}_k$ with Siamese networks --- an online network $W_k^o$ and a target network $W_k^t$; 2) After training, client $k$ uploads $W_k^o$ to the server; 3) The server aggregates them to obtain a new global network $W_g^o$; 4) The server updates $W_k^o$ of client $k$ with $W_g^o$.}
   \label{fig:framework}
   \end{center}
\end{figure*}

\section{An Empirical Study of Federated Self-supervised Learning}
\label{sec:empirical-study}

% In this section, we first propose a generalized framework for federated self-supervised learning that can be used to implement various contrastive learning algorithms. Then, we present empirical study of the performance these algorithms under the framework in federated setting. We end by presenting the impact most important factors for federated self-supervised learning. 

This section first defines the problem and introduces the generalized FedSSL framework. Using the framework, we then conduct empirical studies to reveal deep insights of FedSSL.

\subsection{Problem Definition}

FedSSL aims to learn a generalized representation $W$ from multiple decentralized parties for downstream tasks in the same scenarios. Each party $k$ contains unlabeled data $\mathcal{D}_k = \{\mathcal{X}_k\}$ that cannot be transferred to the server or other parties due to privacy constraints. Data is normally non-IID among decentralized parties \citep{Li2020FedChallenges}; each party could contain only limited data categories (e.g., two out of ten CIFAR-10 classes) \citep{luo2019real-obj-detect}. As a result, each party alone is unable to obtain a good representation \citep{zhuang2021fedu}. The global objective function to learn from multiple parties is
% \begin{equation}
% $\min_{w \in \mathbb{R}^d} f(w) := \sum_{k=1}^K \frac{n_k}{n} f_k(w)$, 
$\min_{w} f(w) := \sum_{k=1}^K \frac{n_k}{n} f_k(w)$, 
% \label{eq:objective}
% \end{equation}
where $K$ is the number of clients, and $n = \sum_{k=1}^K n_k$ is the total data amount. For client $k$, $f_k(w) := \mathbb{E}_{x_k \sim \mathcal{P}_k}[\tilde{f}_k(w;x_k)]$ is the expected loss over data distribution $\mathcal{P}_k$, where $x_k$ is the unlabeled data and $\tilde{f}_k(w;x_k)$ is the loss function. 

% Next, we introduce a generalized FedSSL framework as the first step to address this problem.

% presenting details in each operation as followed.

% The generalized federated self-supervised learning framework aims to enable a wide range of existing contrastive learning methods applicable to learn from decentralized data and possibly extensible to new contrastive learning methods. The framework comprises three key operations: 1) Local Training in clients; 2) Model Aggergation in the server; 3) Model Communication (upload and update) between the server and clients. As illustrated in Figure \ref{fig:framework}, these operations form an end-to-end federated learning pipeline. Firstly, clients download the server initialized model and conduct Local Training with siamese networks (online and target networks) on unlabeled data. Secondly, clients upload the model updates to the server. Thirdly, the server obtains a new global model by aggregating these model updates. Fourthly, the server updates clients' models with the global model. The training iterates these four steps until it meets stopping conditions. We analyze four popular contrastive learning methods, including SimCLR \citep{chen2020simclr}, MoCo (V1 \citep{he2020moco} and V2 \citep{chen2020mocov2}), SimSiam \citep{chen2020simsiam}, and BYOL \citep{grill2020byol}. We summarize their similarities and differences in network designs that affect the three operations.

\subsection{Generalized Framework}
\label{sec:framework}

We introduce a generalized FedSSL framework that empowers existing SSL methods based on Siamese networks to learn from decentralized data under privacy constraints. Figure \ref{fig:framework} depicts the end-to-end training pipeline of the framework. It comprises of three key operations: 1) Local Training in clients; 2) Model Aggregation in the server; 3) Model Communication (upload and update) between the server and clients. We implement and analyze four popular SSL methods --- SimCLR \citep{chen2020simclr}, MoCo (V1 \citep{he2020moco} and V2 \citep{chen2020mocov2}), SimSiam \citep{chen2020simsiam}, and BYOL \citep{grill2020byol}. Variances in Siamese networks of these methods lead to differences in executions in these three operations \footnote{Intuitively, FedSSL is analogous to a SuperClass in OOP, then implementation of each method is a SubClass that inherits FedSSL and overrides methods of local training, model communication, and aggregation.}. 

\textbf{Local Training} \, Firstly, each client $k$ conducts self-supervised training on unlabeled data $\mathcal{D}_k$ based on the same global model $W_g^o$ downloaded from the server. Regardless of SSL methods, clients train with Siamese networks --- an online network $W^o_k$ and a target network $W^t_k$ for $E$ local epochs using cooresponding loss functions $\mathcal{L}$. We classify these SSL methods with two major differences (Figure \ref{fig:ssl-methods} in Appendix \ref{apx:ssl-methods}): 1) Only SimSiam and BYOL contain a \textit{predictor} in the online network, so we denote their online network $W^o_k = (W_k, W_k^p)$, where $W_k$ is the online encoder and $W_k^p$ is the predictor; As for SimCLR and MoCo, $W^o_k = W_k$. 2) SimCLR and SimSiam share identical weights between the online encoder and the target encoder, so $W_k^t = W_k$. In contrast, MoCo and BYOL update the target encoder with EMA of the online encoder in every mini-batch: $W_k^t = m W_k + (1-m) W_k^t$, where $m$ is the momentum value normally set to 0.99. 

\textbf{Model Communication} \, After local training, client $k$ uploads the trained online network $W^o_k$ to the server and updates it with the global model $W_g^o$ after aggregation. Considering the differences of SSL methods, we upload and update encoders and predictors separately: 1) we upload and update the predictor when it presents in local training; 2) we follow the communication protocol \citet{zhuang2021fedu} to upload and update only the online encoder $W_k$ when encoders are different. 

\textbf{Model Aggregation} \, When the server receives online networks from clients, it aggregates them to obtain a new global model $W_g^o = \sum_{k=0}^K \frac{n_k}{n} W_k^o$. $W_g^o = (W_g, W_g^p)$ if predictor presents, otherwise $W_g^o = W_g$, where $W_g$ is the global encoder. Then, the server sent $W_g^o$ to clients to update their online networks. The training iterates these three operations until it meets the stopping conditions. At the end of the training, we use the parameters of $W_g^o$ as the generic representation $W$ for evaluation.

\subsection{Experimental Setup}
\label{sec:experimental-setup}

% We conduct experiments using CIFAR-10, CIFAR-100, and Mini-ImageNet (for transfer learning) datasets. To simulate federated settings, we equally split these datasets to $K$ clients. We simulate non-IID data with label heterogeneity of these datasets, where each client only contains limited classes; We consider number of classes $L = \{2, 4, 6, 8, 10\}$ for CIFAR-10 and $L = \{20, 40, 60, 80, 100\}$ for CIFAR-100. The setting is IID when number of classes $l = 10$ for CIFAR-10 and $l = 100$ for CIFAR-100.

We provide basic experimental setups in this section and describe more details in Appendix \ref{apx:experimental-details}.

\textbf{Datasets} \, We conduct experiments using CIFAR-10 and CIFAR-100 datasets \citep{cifar10-2009}. To simulate federated settings, we equally split a dataset into $K$ clients. We simulate non-IID data with label heterogeneity, where each client contains limited classes --- $l = \{2, 4, 6, 8, 10\}$ number of classes for CIFAR-10 and $l = \{20, 40, 60, 80, 100\}$ for CIFAR-100. The setting is IID when each client contains 10 (100) classes for CIFAR-10 (CIFAR-100).

\textbf{Implementation Details} \, We implement FedSSL in Python using popular deep learning framework PyTorch \citep{paszke2017pytorch}. To simulate federated learning, we train each client on one NVIDIA V100 GPU. These clients communicate with the server through NCCL backend. We use ResNet-18 \citep{he2016resnet} as default network for the encoders and present results of ResNet-50 in Appendix \ref{apx:experimental-results}. The predictor is a two-layer multi-layer perceptron (MLP). By default, we train for $R = 100$ rounds with $K = 5$ clients, $E = 5$ local epoches, batch size $B = 128$, learning rate $\eta = 0.032$ with cosine decay, and non-IID data $l = 2$ ($l=20$) for CIFAR-10 (CIFAR-100).

\textbf{Linear Evaluation} \, We evaluate the quality of representations following linear evaluation \citep{kolesnikov2019revisiting, grill2020byol} protocol. We first learn representations from the FedSSL framework. Then, we train a new linear classifier on the frozen representations. 

% \textbf{Evaluation Protocols} We evaluate the quality of representations following three common evaluation protocols: linear evaluation \citep{} and semi-supervised protocol \citep{}. For all these three protocols, we firstly conduct federated self-supervised learning to obtain a trained model. Using the pre-trained model, three methods evalaute the representation differently: 1) In linear evaluation, we train a new linear classifier on frozen representations; 2) In semi-supervised protocol, we add a new linear classifier and fine-tune the model and the classifier with limited (1\% and 10\%) labeled data.
%  3) In transfer learning, we train the model in one dataset and fine-tune it in another dataset. 

% Based on the generalized framework, we benchmark using these self-supervised learning methods for decentralized data. We split CIFAR-10 and CIFAR-100 datasets \citep{cifar10-2009} into five partitions to simulate five decentralized clients, where each client contains $\frac{1}{5}$ of total classes (2 classes of CIFAR-10 and 20 classes of CIFAR-100) to simulate the non-IID setting. Besides, we use ResNet-18 \citep{he2016resnet} as the backbone and two-layer multi-layer perceptron as the predictor. More experiment settings are described in Section \ref{sec:experiments}. 

\subsection{Algorithm Comparisons}
\label{sec:algorithm-comparison-fedssl}

We benchmark and compare the SSL methods using the FedSSL framework.  To denote the implementation of an SSL method, We add a prefix \textit{Fed} to the name of the SSL method. For example, FedBYOL denotes using BYOL in the FedSSL framework.

Table \ref{tab:ssl-methods-benchmark} compares linear evaluation results of these methods under the non-IID setting of CIFAR datasets. On the one hand, \textit{contrastive} FedSSL methods obtain similar performances. As SimCLR is previously reported to need a large batch size (e.g., $B = 4096$) \citep{chen2020simclr}, it is surprising that FedSimCLR obtains competitive results using the same batch size $B = 128$ as the others. On the other hand, the results of \textit{non-contrastive} FedSSL methods have large variances: FedBYOL achieves the best performance, whereas FedSimSiam yields the worst performance. Since SimSiam is capable to learn as powerful representations as BYOL \citep{chen2020simsiam}, as well as considering that non-contrastive methods are conceptually simpler and more efficient \citep{tian21directpred}, we focus on non-contrastive methods and further investigate the effects of their fundamental components.

% of the models pretrained by these methods. Contrastive methods, including SimCLR and MoCo, obtains similar performances. It is surprising that FedSimCLR, known for requiring large batch size (e.g., 4096), achieves competitive results even with rather small batch size ($B = 128$). Moreover, non-contrastive method, FedBYOL, achieves slightly better results than FedMoCo and FedSimCLR, and greatly outperforms FedSimSiam. Compared with contrastive methods, non-contrastive methods does not rely on memory queue (like MoCo) and is conceptualy simpler. 

% In this paper, we focus on non-contrastive learning methods. SimSiam is report to have similar results as BYOL, but it contradicts our results. To better understand the reasons, we further investigate the impact factors for non-contrastive learning methods under the federated setting.

\def\arraystretch{1.2}

\begin{table}[t]
   \begin{center}
   \caption{Top-1 accuracy comparison of SSL methods using the FedSSL framework on non-IID CIFAR datasets. FedBYOL performs the best, whereas FedSimSiam performs the worst.}
   \label{tab:ssl-methods-benchmark}
   \begin{tabular}{llcc}
   \hline 
   Type & Method & CIFAR-10 (\%) & CIFAR-100 (\%)
   \\
   \Xhline{3\arrayrulewidth}
   % Standalone & 71.98 & 49.69 \\
   {\multirow{3}{*}{Contrastive}} & FedSimCLR & 78.09 $\pm$ 0.14 & 55.58 $\pm$ 0.13 \\ 
   & FedMoCoV1 & 78.21 $\pm$ 0.04 & 56.98 $\pm$ 0.29 \\
   & FedMoCoV2 & 79.14 $\pm$ 0.13 & 57.47 $\pm$ 0.65 \\
   \hline
   {\multirow{2}{*}{Non-contrastive}} & FedSimSiam & 76.27 $\pm$ 0.18 & 48.94 $\pm$ 0.22 \\
   & FedBYOL & \textbf{79.44 $\pm$ 0.99} & \textbf{57.51 $\pm$ 0.09} \\ 
   \hline
   \end{tabular}
   \end{center}
\end{table}

\begin{figure}[t]
   \centering
   \begin{tabular}[c]{ccc}
     \subfloat{\label{fig:predictor}%
         \includegraphics[width=0.33\linewidth]{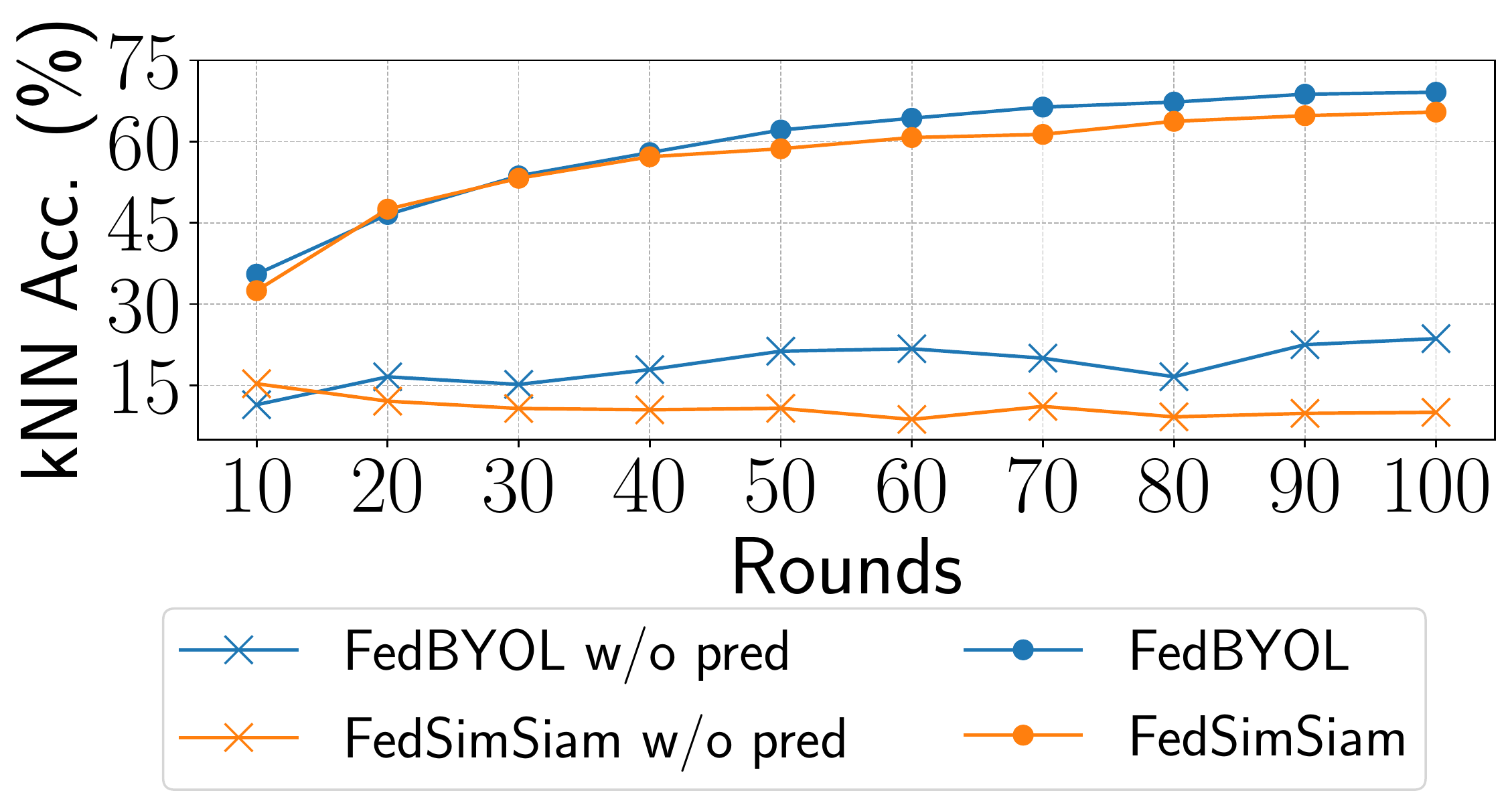}
     }
     \hfill
     \subfloat{\label{fig:stop-grad}%
         \includegraphics[width=0.33\linewidth]{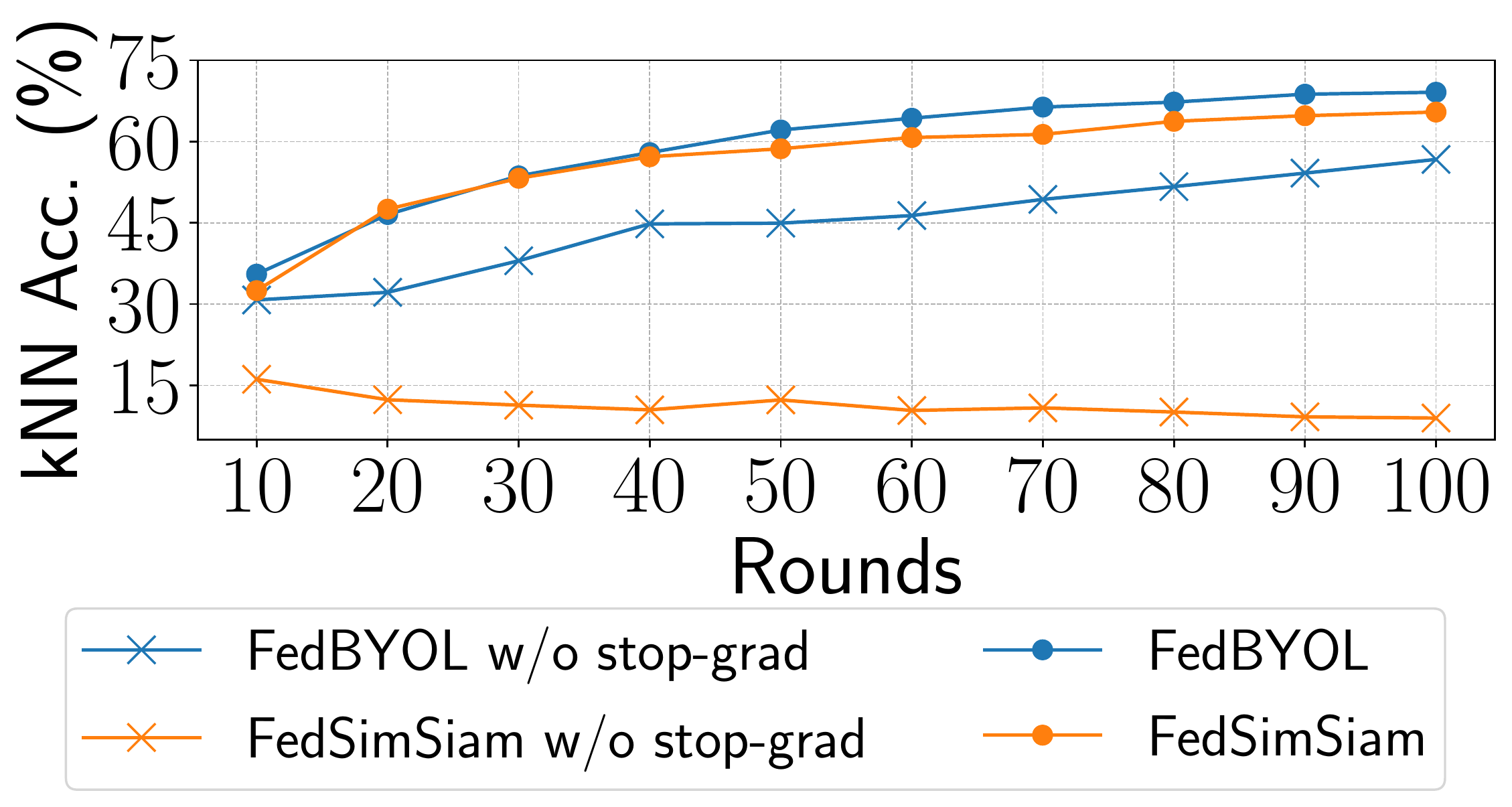}
     }
     \hfill
     \subfloat{\label{tab:stop-grad-t}%
      \begin{tabular}{c|c}
         % \hline
         FedBYOL &  acc (\%)
         \\
         \hline
         % \Xhline{3\arrayrulewidth}
         % FedSimSiam w/o stop-grad & 11.62 \\ 
         % FedSimSiam & 74.91 \\
         % \hline
         w/o stop-grad & 75.73 \\
         w/ stop-grad & 79.44 \\
         % \hline
      \end{tabular}
     }
   \end{tabular}
   \caption{Comparison of non-contrastive FedSSL methods with and without (w/o) predictor (pred) or stop-gradient (stop-grad) on non-IID CIFAR-10 dataset. Without predictor, both FedSimSiam and FedBYOL drops performance on kNN testing accuracy (left plot). Without stop-gradient, FedBYOL retains competitive results on kNN testing accuracy (middle plot) and linear evaluation (right table).}
   \label{fig:pred-and-stop-grad}
\end{figure}

\subsection{Impact of Factors of Non-contrastive Methods}

This section analyzes the impact of fundamental components of non-contrastive FedSSL methods. From empirical studies, we obtain the following insights: 1) predictor is essential; 2) EMA and stop-gradient improves performances; 3) Local encoders should retain local knowledge of the non-IID data; 4) Target encoder should gain knowledge from the online encoder. Details are as followed.

\textbf{Predictor} is essential. Figure \ref{fig:pred-and-stop-grad} (left plot) presents the kNN testing accuracy as a monitoring process for FedBYOL and FedSimSiam with and without predictors. Without predictors, both methods can barely learn due to collapse in local training. It affirms the vital role of predictor \citep{chen2020simsiam,tian21directpred} even when learning from decentralized data. 

\textbf{Stop-gradient} operation is previously indicated as an essential component for SimSiam and BYOL \citep{tian21directpred}, but it is \textit{not essential} for FedBYOL. Stop-gradient prevents stochastic gradient optimization on the target network. Figure \ref{fig:pred-and-stop-grad} shows that FedSimSiam without stop-gradient collapses, whereas FedBYOL without stop-gradient still achieves competitive performance. It is because online and target encoders are significantly different in FedBYOL as the online encoder is updated by the global encoder every communication round. In contrast, SimSiam or FedSiamSiam share weights between online and target encoders, so removing stop-gradient leads to collapse.

\textbf{Exponential Moving Average (EMA)} is not essential, but it helps improve performance. Table \ref{tab:update-both} (first and second rows) shows that FedBYOL outperforms FedSimSiam at different levels of non-IID data, which is represented by \{2, 4, 6, 8, 10\} classes per client of CIFAR-10. EMA is the main difference between SimSiam and BYOL, indicating that EMA is helpful to improve performance. Based on these results, we further analyze the underlying impact of EMA on the encoders below.

\begin{table}[t]
   \begin{center}
   \begin{tabular}{lccccc}
      \hline
      \multicolumn{1}{l}{\multirow{2}{*}{Method}} &
      \multicolumn{5}{c}{\# of classes per client (\%)}
      \\ 
      \cline{2-6}
      & 2 & 4 & 6 & 8 & 10 (iid)
      \\
      % \# of classes per client & 2 & 4 & 6 & 8 & 10 (iid)
      \Xhline{3\arrayrulewidth}
      FedBYOL & 79.44 & 82.82 & 83.02 & 84.57 & 84.20 \\
      FedSimSiam & 76.27 & 79.34 & 80.17 & 80.92 & 80.50 \\ 
      FedBYOL, update-both & 74.50 & 78.77 & 83.02 & 84.56 & 83.80 \\
      \hline
   \end{tabular}
   \caption{Top-1 accuracy comparison on various non-IID levels --- the number of classes per client on the CIFAR-10 dataset. Update-both means updating both $W_k$ and $W_k^t$ with $W_g$.}
   \label{tab:update-both}
   \end{center}
\end{table}

\begin{figure}[t]
   \centering
   \begin{tabular}[c]{cc}
     \subfloat{\label{fig:stop-grad}%
         \includegraphics[width=0.44\linewidth]{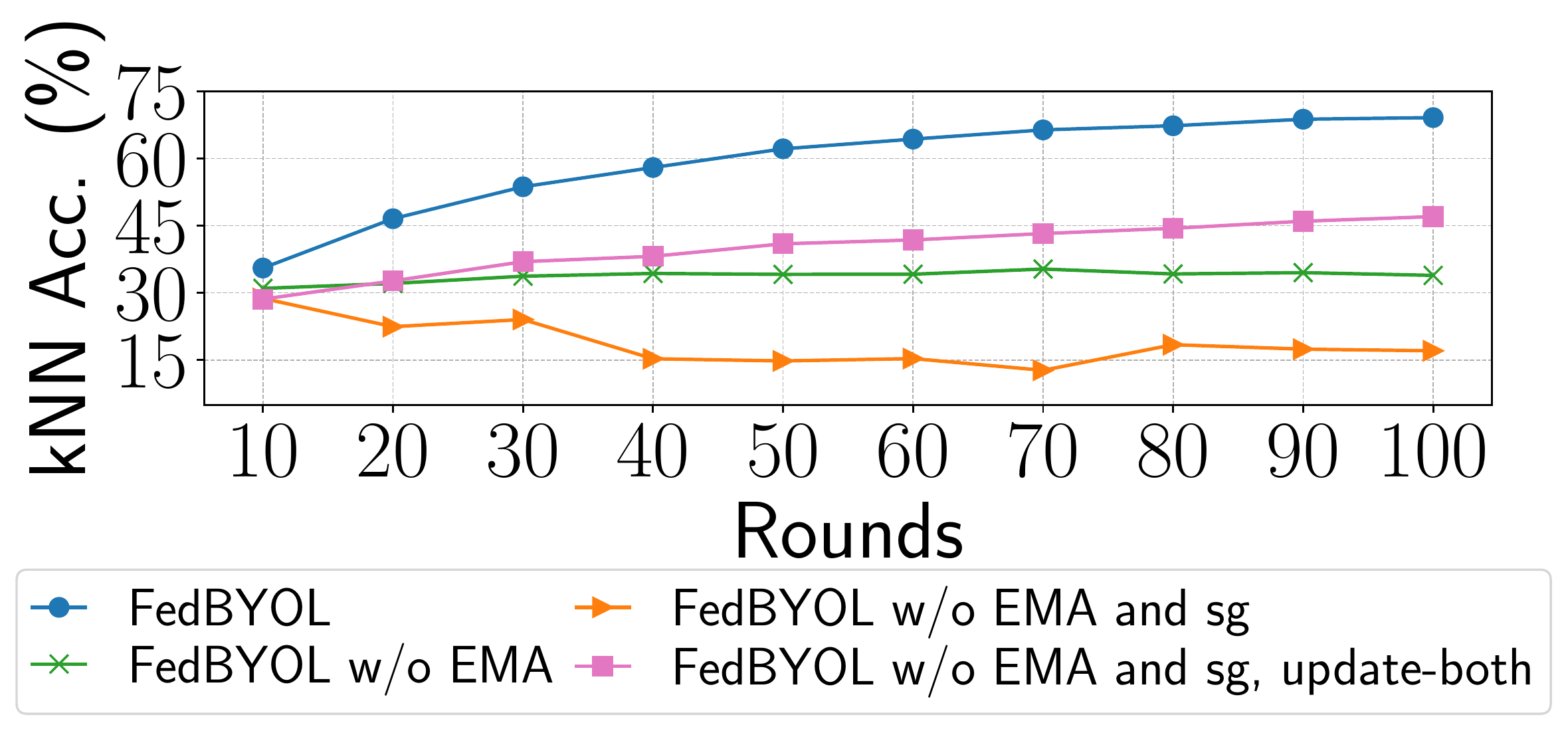}
     }
     \hfill
     \subfloat{\label{tab:stop-grad}%
     \begin{tabular}{c|c}
         % FedBYOL & CIFAR-10 & CIFAR-100
         FedBYOL & acc (\%)
         \\
         \hline 
         % SimSiam original & $\checkmark$ & - & 80.50 & 74.91 & & 50.28 & 49.95 \\
         % SimSiam no sg & - & - & 12.00 & 11.62 & & 1.90 & 2.00 \\ 
         % FedBYOL & \textbf{79.44} & \textbf{57.56} \\
         % BYOL no sg & - & $\checkmark$ & 80.45 & 75.73 & & 51.17 & 47.06 \\
         % w/o EMA & 50.20 & 25.60 \\
         % w/o EMA and stop-grad & 11.97 & 2.79\\
         % w/o EMA and stop-grad, update both & 68.75 & 41.91 \\
         w/o EMA & 50.20 \\
         w/o EMA and stop-grad & 11.97 \\
         w/o EMA and stop-grad, update-both & 68.75 \\
         original & 79.44 \\
      \end{tabular}
    }
   \end{tabular}
   \caption{Comparison of FedBYOL without exponential moving average (EMA) and stop-gradient (sg) on the non-IID CIFAR-10 dataset. FedBYOL w/o EMA and sg can hardly learn, but updating both $W_k$ and $W_k^t$ with $W_g$ (update-both) enables it to achieve comparable results.}
   \label{fig:no_ema_no_sg}
 \end{figure}

\textbf{Encoders} that retain local knowledge of non-IID data helps improve performance. EMA in FedBYOL allows the parameters of the online encoder to be different from the target encoder. As a result, the global encoder only updates the online encoder, not the target encoder. We hypothesize that retaining such local knowledge of data in the target encoder is beneficial especially when the data distribution is highly skewed. For comparison, we remove such local knowledge by updating both online and target encoders with the global model. 
Table \ref{tab:update-both} shows that FedBYOL with both encoders updated leads to lower performance than FedBYOL; It achieves results close to FedSimSiam when the data distribution is more skewed (2 or 4 classes per client). These results demonstrate the importance of keeping local knowledge in the encoders. Besides, the results of \{6, 8, 10\} classes per client also indicate the benefit of EMA.

\textbf{Target encoder} is essential to gain knowledge from the online encoder. Figure \ref{fig:no_ema_no_sg} shows that FedBYOL without EMA can merely learn, and FedBYOL without EMA and stop-gradient (sg) degrades in performance. In both cases, the target encoder is either never updated (w/o EMA) or is updated only through backpropagation (w/o EMA and sg) --- not updated by the online encoder. On the other hand, we also identify that FedSSL methods in Table \ref{tab:ssl-methods-benchmark}, which achieve competitive results, all update target encoders with knowledge of online encoders (the global encoder is the aggregation of the online encoder). We argue that target encoder is crucial to gain knowledge from the online encoder to provide contrastive targets. We further validate it by using the global encoder to update both online and target encoders when removing both EMA and stop-gradient. Figure \ref{fig:no_ema_no_sg} shows that such method improves performance and achieves comparable results.

\section{Divergence-aware Dynamic Moving Average Update}
\label{sec:fedema}

Built on the FedSSL framework, we propose Federated Divergence-aware EMA update (FedEMA) to further mitigate non-IID data challenges. Since FedBYOL contains all components that help improve performance, we adopt it as the baseline and optimize the \textit{model update} operation. 

Non-IID data causes the global model to diverge from centralized training \citep{zhuang2021fedu}. Inspired by the insight that retaining local knowledge of non-IID data helps improve performance, we propose to update the online network via EMA of the global network. Compared with FedBYOL that replaces the online network with the global network, FedEMA fuses local and global knowledge effectively through EMA update, where the decay rate of EMA is dynamically measured by model divergences. Figure \ref{fig:dynamic_ema} depicts our proposed FedEMA method. The formulation is as followed:

\begin{equation}
   W_k^{r} = \mu W_k^{r-1} + (1 - \mu) W_g^{r}
\label{eq:encoder},
\end{equation}

\begin{equation}
   W_k^{p,r} = \mu W_k^{p,r-1} + (1 - \mu) W_g^{p,r}
\label{eq:predictor},
\end{equation}

\begin{equation}
   \mu = \text{min}(\lambda \norm{W_g^{r} - W_k^{r-1}}, 1),
   \label{eq:mu}
\end{equation}

where $W_k^r$ and $W_k^{p,r}$ are the online encoder and predictor of client $k$ at training round $r$; $W_g^{r}$ and $W_g^{p,r}$ are the global encoder and predictor; $\mu$ is the decay rate, measured by the divergence between global and online encoders; $\lambda$ is a scaler to adjust the level of model divergence, which is measured by calculating the $l_2$-norm of the global and online encoders. We summarize FedEMA in Algorithm \ref{algo:fedema}. FedEMA can be regarded as a generalization of FedBYOL --- they are the same when $\lambda = 0$.

Scaler $\lambda$ plays a vital role to adapt FedEMA for different levels of divergence caused by the data. The divergence between global and online encoders varies when the settings of federated learning change. For example, different degrees of non-IID settings would result in different divergences.
Since characteristics of data are unknown before training as they are unlabeled, we propose a practical \textit{autoscaler} to calculate a personalized $\lambda_k$ for each client $k$ automatically. The formula is $\lambda_k = \frac{\tau}{||W_g^{r+1} - W_k^r||}$, where $\tau \in [0, 1]$ is the expected value of $\mu$ at round $r$. We calculate $\lambda_k$ only once at the earliest round $r$ that client $k$ is sampled for training. When the same set of clients are sampled for training, $\lambda_k = \frac{\tau}{||W_g^1 - W_k^0||}$ is calculated at round $r=1$.

% more local epochs in local training would intensify the impact of non-IID data, whereas a smaller local epoch reduces the divergence. 

\begin{minipage}[t]{\textwidth}
   \begin{minipage}{0.46\textwidth}
      \begin{algorithm}[H]
         \centering
         \caption{Our proposed FedEMA}\label{algo:fedema}
         \begin{algorithmic}[1]
           \State \text{\textbf{\underline{ServerExecution:}}}
           \State Init $W_g^0$ and $W_g^{p,0}$, init $\lambda_k$ to be \textit{null}
           \For{\textit{each round} $r = 0, 1, ..., R$}
              \State $S_t \gets$ (Selection of K clients)\;
              \For{\textit{client} $k \in S_t$ \textit{in parallel}}
                  \State $W_k^r,W_k^{p,r}\gets$Client($W_g^r$,$W_g^{p,r}$,$r$,$\lambda_k$)\;
              \EndFor
              % \State $(W_g^{r+1}, W_g^{p,r+1}) \gets \sum_{k \in S_t} \frac{n_k}{n} (W_k^r, W_k^{p,r})$
              \State $W_g^{r+1} \gets \sum_{k \in S_t} \frac{n_k}{n} W_k^r$
              \State $W_g^{p,r+1} \gets \sum_{k \in S_t} \frac{n_k}{n} W_k^{p,r}$
      
              \For{\textit{client} $k \in S_t$}
              \State \textcolor{purple}{$\lambda_k \gets \frac{\tau}{||W_g^{r+1} - W_k^{r}||}$} \textbf{if} $\lambda_k$ is \textit{null}
               % \State \textbf{if} $\lambda_k$ is \textit{null} \textcolor{purple}{$\lambda_k \gets \frac{\tau}{||W_g^{r+1} - W_k^{r}||}$}
               \EndFor
            %   \If{$r == 1$} 
            %    \EndIf
           \EndFor
           \State \textbf{Return}  $W_g^{R}$
         %   \State 
           \State \underline{\textbf{Client}} ($W_g^r$, $W_g^{p,r}$, $r$, $\lambda_k$):
           \If{$\lambda_k$ is \textit{null} \textit{or} not selected in $r-1$}
               % \State \textcolor{purple}{$\lambda_k \gets \frac{\tau}{\norm{W_g^1 - W_k^0}}$}
               \State $W_k, W_k^t, W_k^{p} \gets W_g^{r}, W_g^{r}, W_g^{p,r}$
               % \State $W_k^o, W_k^t \gets W_g^{o,r}, W_g^{r}$; ($W_k, W_k^{p}) \gets W_k^{o,r}$
               % \State \textcolor{purple}{}
            \Else
               % \State \textcolor{purple}{$\lambda_k = \frac{\tau}{\norm{W_g^1 - W_k^0}}$}
               \State \textcolor{purple}{$\mu \gets \text{min}(\lambda_k \norm{W_g^r - W_k^{r-1}}, 1)$}
               \State \textcolor{purple}{$W_k \gets \mu W_k^{r-1} + (1 - \mu) W_g^{r}$}
               \State \textcolor{purple}{$W_k^{p} \gets \mu W_k^{p,r-1} + (1 - \mu) W_g^{p,r}$}
           \EndIf
           
          \For{\textit{local epoch} $e = 0, 1, ..., E-1$}
              \For{$b \in \mathcal{B}$ \textit{data batches with size} $B$}
                  \State $W_k^o \leftarrow W_k^o - \eta \triangledown \mathcal{L}_{W_k^o, W_k^t}(W_k^o; b)$ 
                 %  \tcp*{$W_k^o = (W_k, W_k^p)$}
                  \State $W_k^t \leftarrow mW_k^t + (1-m)W_k$
              \EndFor
          \EndFor
          \State \textbf{Return} $W_k^r$, $W_k^{p,r}$
         \end{algorithmic}
      \end{algorithm}
   \end{minipage}
   \hfill
   \begin{minipage}{0.5\textwidth}
       \centering
         \includegraphics[width=1\linewidth]{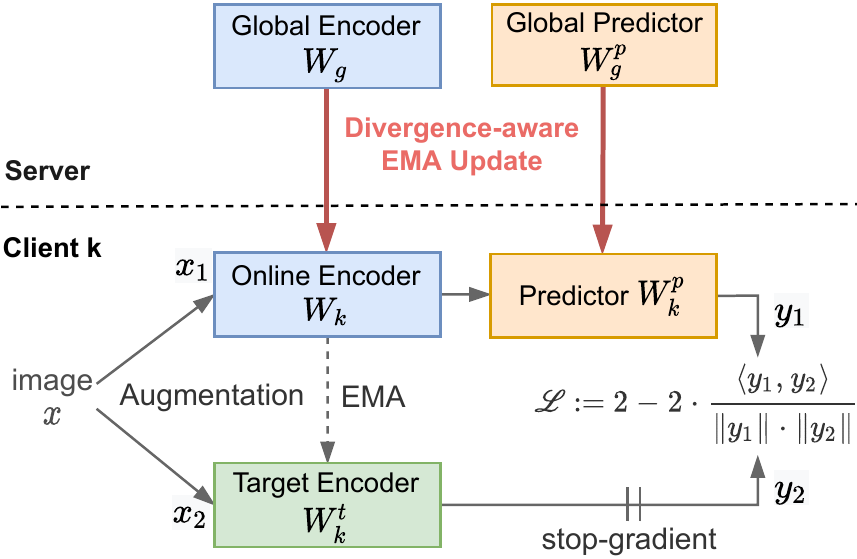}
         \captionof{figure}{Illustration of our proposed Federated Divergence-aware Exponential Moving Average update (FedEMA). Compared with FedBYOL that simply updates the online network $W_k^o$ of client $k$ with the global network $W_g^o$, we propose to update them via EMA of the global network following Eqn \ref{eq:encoder} and \ref{eq:predictor}, where the decay rate $\mu$ is dynamically measured the divergences between the online encoder $W_k$ and the global encoder $W_g$ (Eqn \ref{eq:mu}). The online network, $W_k^o = (W_k, W_k^p)$, is the concatenation of the online encoder $W_k$ and the predictor $W_k^p$.}
         \label{fig:dynamic_ema}
   \end{minipage}
\end{minipage}

The intuition of FedEMA is to retain more local knowledge when divergence is large and incorporate more global knowledge when divergence is small. When model divergence is large, keeping more local knowledge is more beneficial for the non-IID data. Since the global network is the aggregation of online networks, representing global knowledge from clients. When divergence is small, adapting more global knowledge help improve model generalization. Since model divergence is larger at the start of training (Figure \ref{fig:divergence_analysis}), it is practical to choose larger $\tau \in [0.5, 1)$; $\tau = 1$ is not considered because only local knowledge is used when $\tau = 1$. We use $\tau = 0.7$ by default in experiments.

\section{Evaluation}
\label{sec:experiments}

This section follows the experimental setup in Section \ref{sec:experimental-setup} to evaluate FedEMA in the linear evaluation and semi-supervised learning. We also provide ablation studies of important hyperparameters.

\begingroup
\setlength{\tabcolsep}{0.5em}
\begin{table*}[t]
   \begin{center}
   \begin{tabular}{lccccc}
   \hline
   \multicolumn{1}{l}{\multirow{2}{*}{Method}} &
   \multicolumn{2}{c}{CIFAR-10 (\%)} &
   \multicolumn{1}{c}{} &
   \multicolumn{2}{c}{CIFAR-100 (\%)}
   \\ 
   \cline{2-3} \cline{5-6}
   \multicolumn{1}{c}{} & IID & Non-IID & & IID & Non-IID
   \\
   % \hline 
   \Xhline{3\arrayrulewidth}
   Standalone training & 82.42 $\pm$ 0.32 & 74.95 $\pm$ 0.66 & & 53.88 $\pm$ 2.24 & 52.37 $\pm$ 0.93 \\
   FedBYOL & 84.29 $\pm$ 0.18 & 79.44 $\pm$ 0.99 & & 54.24 $\pm$ 0.24 & 57.51 $\pm$ 0.09 \\ 
   FedU \citep{zhuang2021fedu} & 83.96 $\pm$ 0.18 & 80.52 $\pm$ 0.21 & & 54.82 $\pm$ 0.67 & 57.21 $\pm$ 0.31 \\ 
   FedEMA ($\lambda=0.8$) & \textbf{85.59 $\pm$ 0.25} & \textbf{82.77 $\pm$ 0.08} & & \textbf{57.86 $\pm$ 0.15} & \textbf{61.21 $\pm$ 0.54} \\ 
   % FedEMA (autoscaler, $\tau=0.8$) & \textbf{85.58 $\pm$ 0.13} & \textbf{83.36 $\pm$ 0.17} & & \textbf{57.29 $\pm$ 0.11} & \textbf{60.91 $\pm$ 0.26} \\ 
   FedEMA (autoscaler, $\tau=0.7$) & \textbf{86.26 $\pm$ 0.26} & \textbf{83.34 $\pm$ 0.39} & & \textbf{58.55 $\pm$ 0.34} & \textbf{61.78 $\pm$ 0.14} \\
   \hline
   % \multicolumn{6}{l}{\textit{Upper-bound methods: centralized unsupervised learning and supervised federated learning}} \\
   % \hline
   % FedAvg (Supervised) & 91.97 & 74.21 & & 65.76 & 64.06 \\
   BYOL (Centralized) & 90.46 $\pm$ 0.34 & - & & 65.54 $\pm$ 0.47 & - \\
   % FedAvg (Supervised) & ResNet-50 & 23M & 91.51 & 67.74 & & 65.77 & 64.38 \\
   % BYOL (Centralized) & ResNet-50 & 23M & 91.85 & - & & 66.51 & - \\
   
   \hline
   \end{tabular}
   \caption{Top-1 accuracy comparison under linear probing on CIFAR datasets. Our proposed FedEMA outperforms all other methods. Full results are in Table \ref{tab:linear-eval-full}.}
   \label{tab:linear-eval}
   \end{center}
\end{table*}    
\endgroup

\begingroup
\setlength{\tabcolsep}{0.5em}
\begin{table*}[t]
   \begin{center}
   \begin{tabular}{lcccccccc}
   \hline
   \multicolumn{1}{l}{\multirow{2}{*}{Method}} &
   \multicolumn{2}{c}{CIFAR-10 (\%)} &
   \multicolumn{1}{c}{} &
   \multicolumn{2}{c}{CIFAR-100 (\%)}
   \\ 
   \cline{2-3} \cline{5-6}
   \multicolumn{1}{c}{} & 1\% & 10\% & & 1\% & 10\%
   \\
   \Xhline{3\arrayrulewidth}

   Standalone training & 61.37 $\pm$ 0.13 & 69.06 $\pm$ 0.24 & & 21.37 $\pm$ 0.73 & 39.99 $\pm$ 0.87 \\
   FedBYOL & 70.48 $\pm$ 0.30 & 76.95 $\pm$ 0.46 & & 30.21 $\pm$ 0.40 & 47.07 $\pm$ 0.14 \\
   FedU \citep{zhuang2021fedu} & 69.52 $\pm$ 0.73 & 77.06 $\pm$ 0.55 & & 29.00 $\pm$ 0.27 & 46.67 $\pm$ 0.06 \\
   FedEMA ($\lambda=1$) & \textbf{72.78 $\pm$ 0.66} & \textbf{79.01 $\pm$ 0.30} & & \textbf{32.49 $\pm$ 0.22} & \textbf{49.82 $\pm$ 0.36} \\
   % FedEMA (autoscaler, $\tau=0.8$) & \textbf{73.50 $\pm$ 0.64} & \textbf{79.80 $\pm$ 0.39} & & \textbf{32.91 $\pm$ 0.39} & \textbf{50.31 $\pm$ 0.42} \\
   FedEMA (autoscaler, $\tau=0.7$) & \textbf{73.44 $\pm$ 0.22} & \textbf{79.49 $\pm$ 0.34} & & \textbf{33.04 $\pm$ 0.23} & \textbf{50.48 $\pm$ 0.11} \\

   \hline
   BYOL (Centralized) & 87.67	$\pm$ 0.15 & 87.89 $\pm$ 0.05 & & 40.96 $\pm$ 0.58 & 56.60 $\pm$ 0.33 \\
   \hline
   \end{tabular}
   \caption{Top-1 accuracy comparison on 1\% and 10\% of labeled data for semi-supervised learning on non-IID CIFAR datasets. FedEMA outperforms other methods. Full results are in Table \ref{tab:semi-sup-full}.}
   \label{tab:semi-sup}
   \end{center}
\end{table*}    
\endgroup

\subsection{Algorithm Comparisons}
\label{sec:algorithm-comparison-fedema}

To demonstrate the effectiveness of FedEMA, we compare it with the following methods: 1) standalone training, where a client learns independently using BYOL; 2) FedCA, which is proposed in \citet{zhang2020fedca}; 3) FedBYOL as the baseline; 4) FedU, which is proposed in \citep{zhuang2021fedu}. Besides, we also present results of possible upper bounds that learn representations with centralized data using BYOL.

\textbf{Linear Evaluation} \, Table \ref{tab:linear-eval} shows that FedEMA outperforms other methods on different settings of CIFAR datasets. Specifically, the performance is more 3\% higher than existing methods in most settings. Besides, our proposed autoscaler achieves similar results as $\lambda = 0.8$. More experiments on larger number of clients $K$ and random selection of clients are provided in Table \ref{tab:scalability} in Appendix \ref{apx:experimental-results}.

% We run these experiments with $\lambda = 0.8$, while FedEMA still achieves 82.64\% on CIFAR-10 and 60.7\% on CIFAR-100 on non-IID settings when $\lambda=1$. 

% Besides, the results on IID settings are close to the possible upper bounds. 
% Specifically, the performance is more than 4\% higher than the SOTA on non-IID of CIFAR-10. We run CIFAR-10 datasets experiments with $\lambda = 0.8$; The performance of non-IID still achieves 82.54\% when $\lambda=1$.

\textbf{Semi-supervised Learning} \, We also assess the quality of representations following the semi-supervised learning protocol \citep{zhai2019s4l, chen2020simclr} --- we add a new two-layer MLP on top of the encoder and fine-tune the whole model with limited (1\% and 10\%) labeled data for 100 epochs. Table \ref{tab:semi-sup} indicates that FedEMA consistently outperforms other methods on non-IID settings of CIFAR datasets and our autoscaler outperforms manual-selected $\lambda = 1$.

\begingroup
\floatsetup[subfigure]{captionskip=-10pt}%
\setlength{\tabcolsep}{0.3em}
\begin{figure}[t!]
   \setlength\fboxsep{0pt}\setlength\fboxrule{0.75pt}
   \begin{floatrow}
   \ffigbox[0.65\textwidth]{
       \begin{subfloatrow}[2]
       %\fbox{
       \ffigbox[.35\textwidth]{
           \begin{tabular}{c|c}
               Method & acc (\%) \\
               \Xhline{3\arrayrulewidth}
               FedBYOL & 79.44 \\ 
               FedEMA predictor only & 81.13 \\ 
               FedEMA encoder only & 82.39 \\ 
               FedEMA (ours)& 82.77 \\ 
           \end{tabular}
       }{\subcaption*{}}
       
       \ffigbox[.3\textwidth]{
           \includegraphics[width=0.9\linewidth]{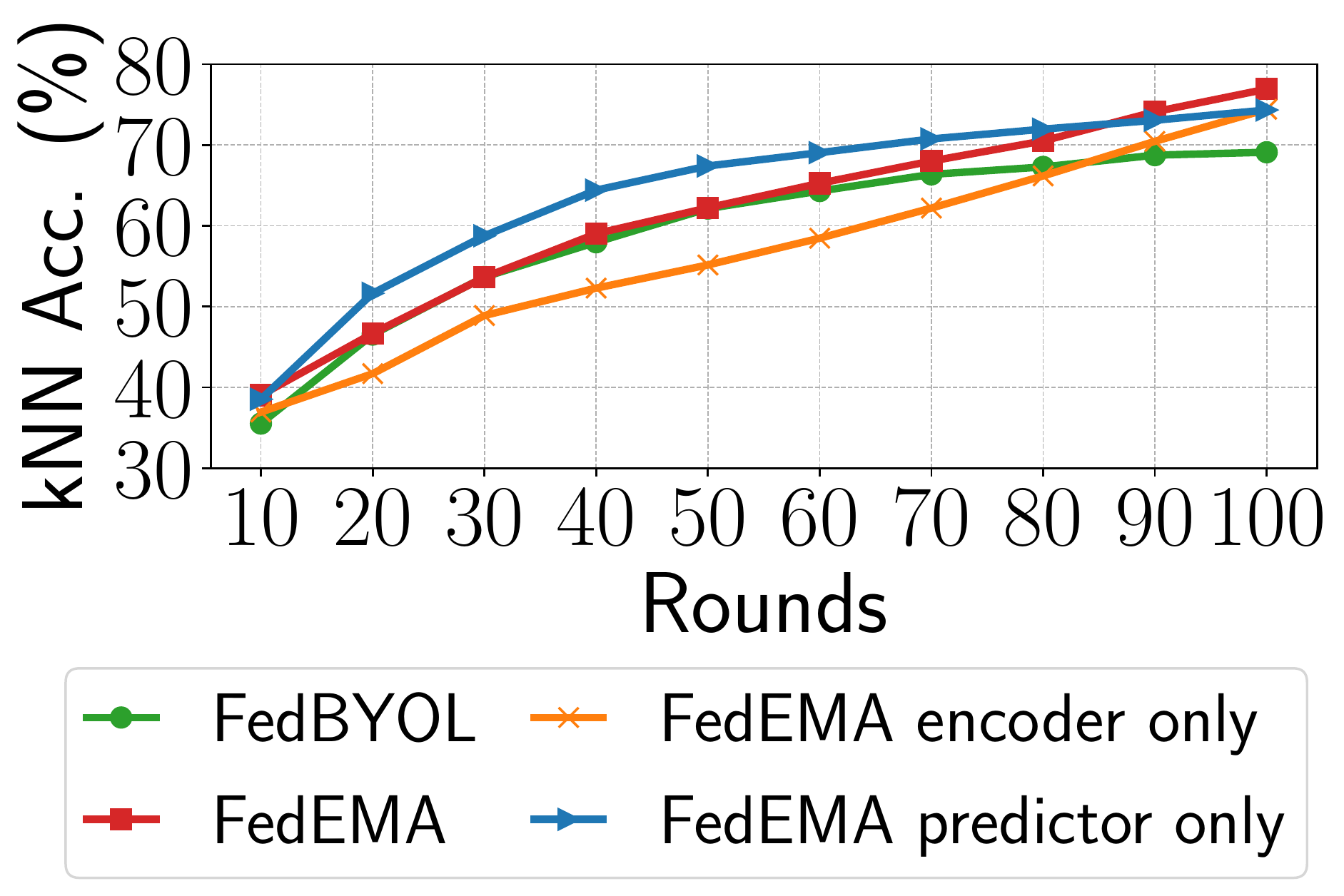}
       }{\subcaption*{}}
       \end{subfloatrow}%
   }{
     \caption{Ablation studies of FedEMA: applying EMA on either predictor or encoder leads to better performance on CIFAR-10.}
     \label{fig:ablation}
   }
   \hfill
   \ffigbox[.3\textwidth]{
       \caption{Changes of divergence throughout training. }
       \label{fig:divergence_analysis}
     }{
       \includegraphics[width=1\linewidth]{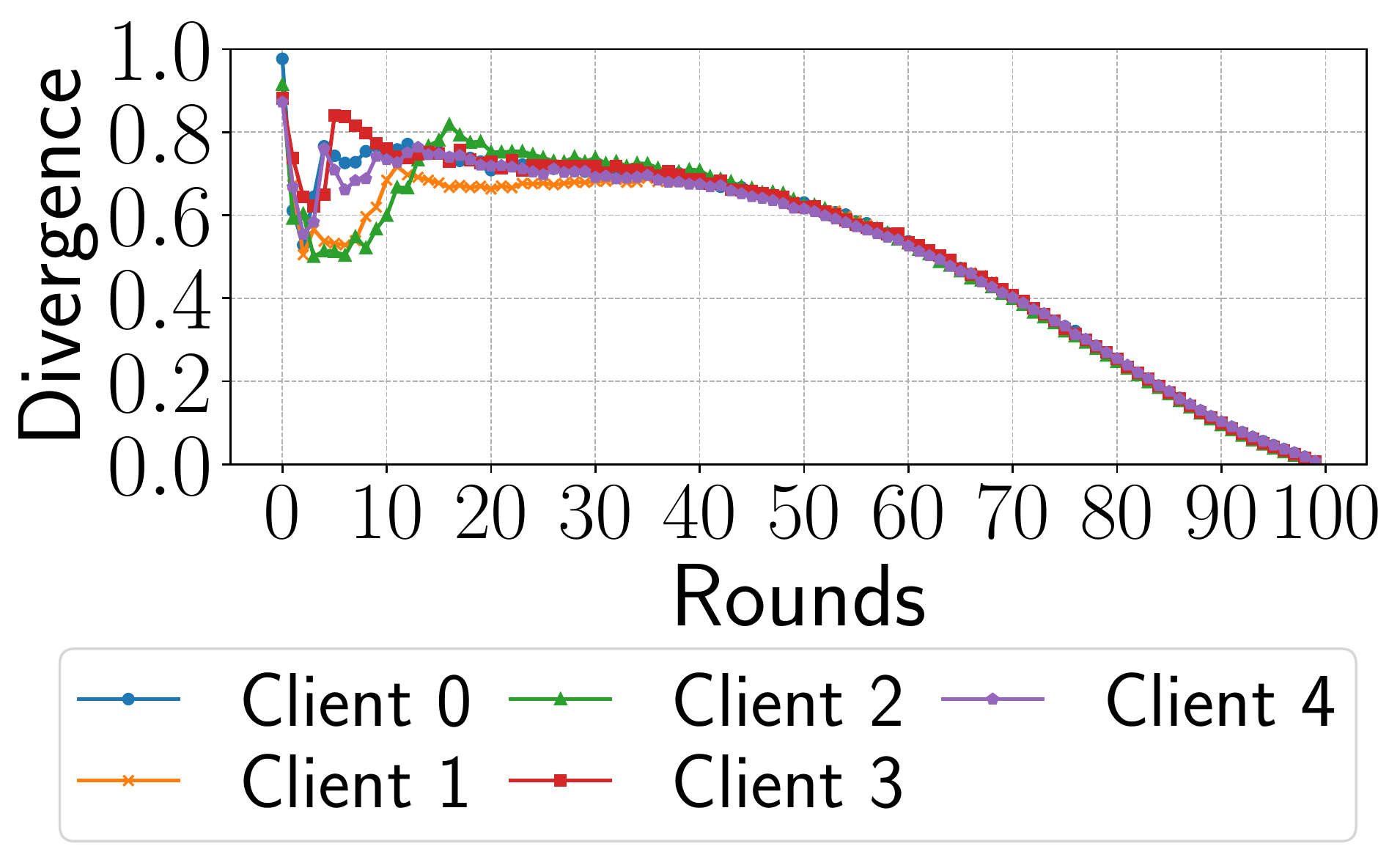}
     }
   \end{floatrow}
\end{figure}%
\endgroup

\subsection{Ablation Studies} 

\textbf{Ablation on FedEMA} \, We analyze whether we need to update both online encoder (Eqn \ref{eq:encoder}) and predictor (Eqn \ref{eq:predictor}) in FedEMA. Figure \ref{fig:ablation} shows that updating only the encoder or predictor leads to better performance; only updating predictor also leads to faster convergence. Their combination results in the best performance. These results demonstrate the effectiveness of updating both predictor and encoder in FedEMA. More results on other settings are provided in Table \ref{tab:linear-eval-full} in Appendix \ref{apx:experimental-results}.

\textbf{Changes of Divergence} \,  Figure \ref{fig:divergence_analysis} illustrates that the divergence between global encoder and online encoder (Eqn \ref{eq:mu}) decreases gradually as training proceeds. It validates our intuition that more local knowledge is used at the start of training when divergence is larger. Besides, clients can update at their own pace depending on the divergence caused by their local dataset.

\begin{figure}[t!]
   \centering
   \begin{subfigure}[t]{0.32\textwidth}
       \includegraphics[width=\textwidth]{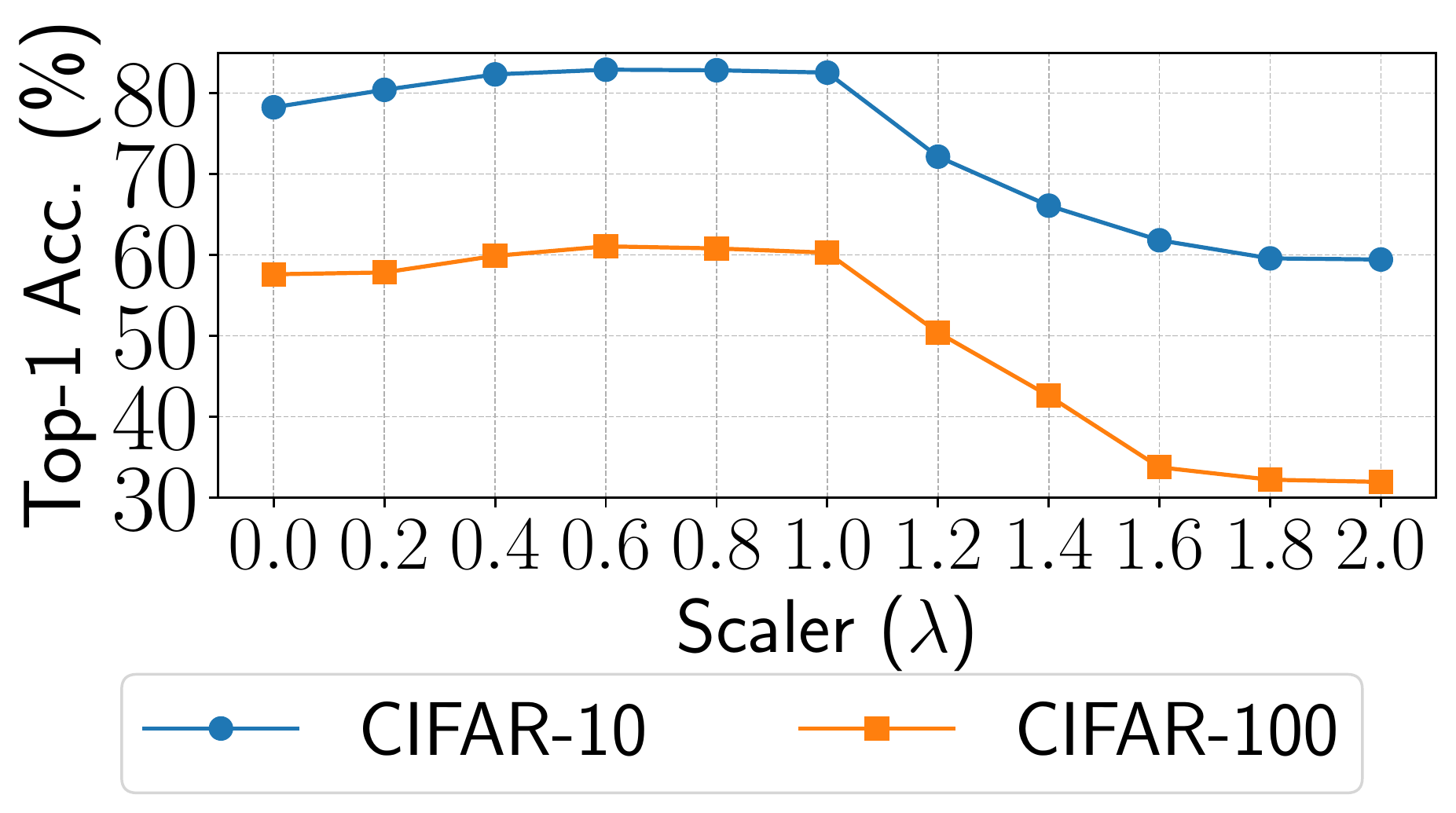}
       \caption{Scaler $\lambda$}
       \label{fig:scaler}
   \end{subfigure}
   \hfill
   \begin{subfigure}[t]{0.32\textwidth}
      \includegraphics[width=\textwidth]{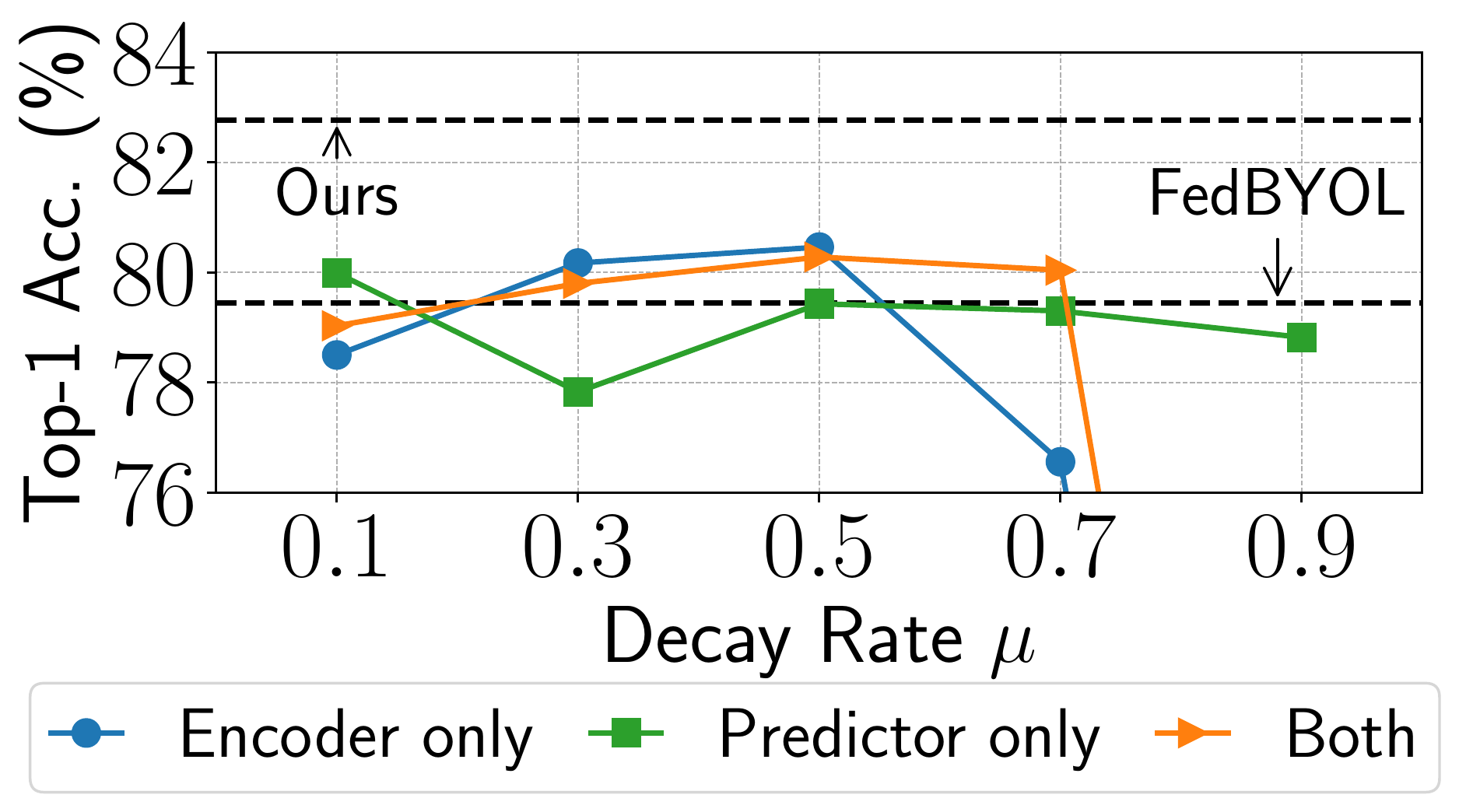}
      \caption{Constant $\mu$}
      \label{fig:const-ema}
  \end{subfigure}
  \hfill
  \begin{subfigure}[t]{0.32\textwidth}
   \includegraphics[width=\textwidth]{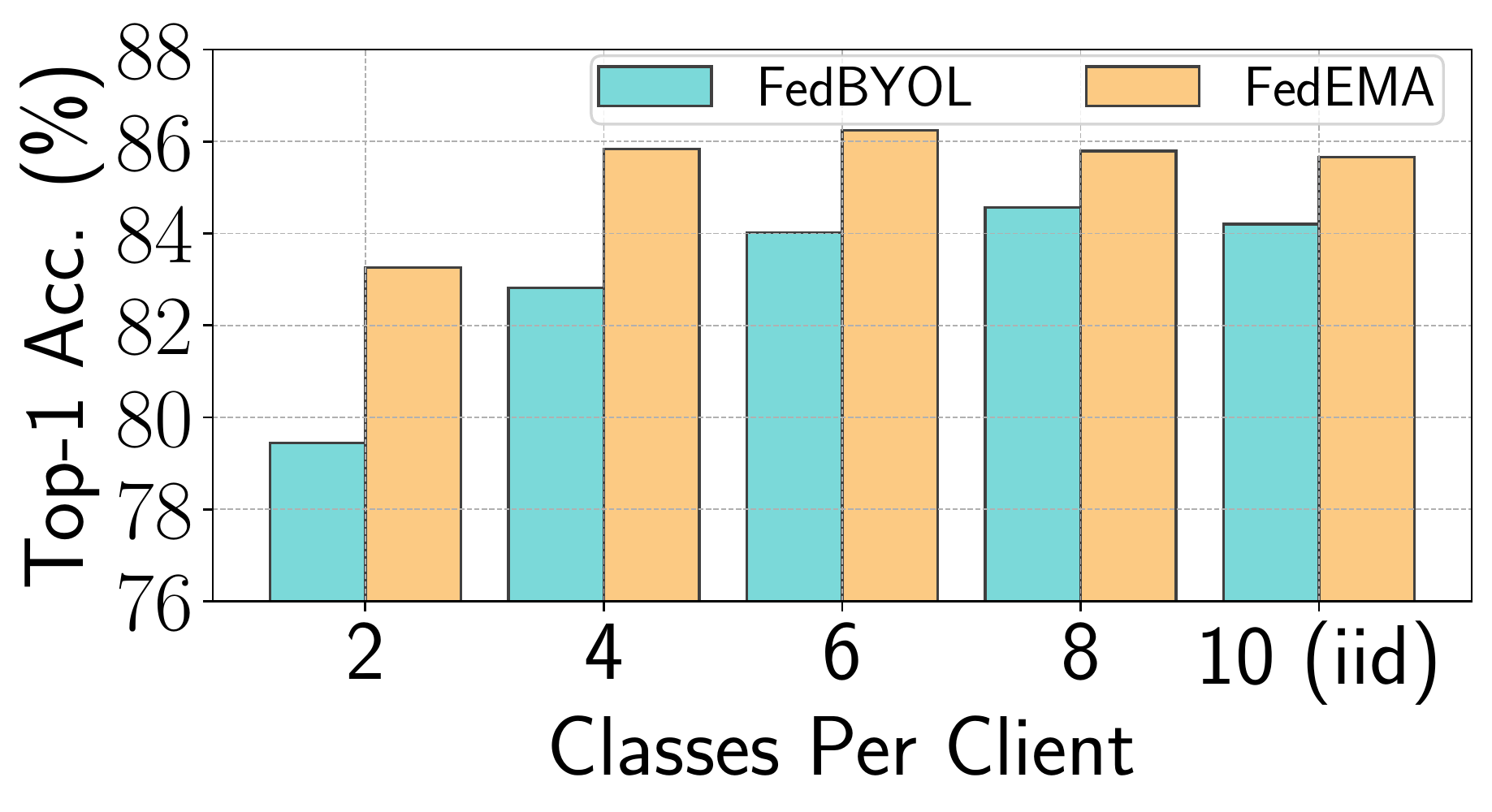}
   \caption{Non-IID level}
   \label{fig:non-iid-level}
   \end{subfigure}
   % \begin{subfigure}[t]{0.32\textwidth}
   %     \includegraphics[width=\textwidth]{charts/ema_predictor_encoder.pdf}
   %     \caption{Constant $\mu$ on encoder or predictor}
   %     \label{fig:const-ema}
   % \end{subfigure}
   % \hfill
%    \begin{subfigure}[t]{0.32\textwidth}
%       \includegraphics[width=\textwidth]{charts/ema_encoder_predictor.pdf}
%       \caption{Constant $\mu$ on encoder and predictor}
%       \label{fig:ema-encoder-predictor}
%   \end{subfigure}
  \caption{Ablation study on scaler $\lambda$, decay rate $\mu$, and non-IID levels of the CIFAR-10 dataset: (a) analyzes the impact of scaler $\lambda$ on performance; (b) compares using constant $\mu$ on encoder, predictor, or both; (c) studies the impact of different non-IID levels.}
  \label{fig:static-ema}
\end{figure}

\textbf{Scaler $\lambda$} \, We study the impact of $\lambda$ with values in $[0, 2]$ with interval of 0.2 in Figure \ref{fig:scaler}. $\lambda > 1$ leads to a significant performance drop because it results in $\mu = 1$ at the start of training on CIFAR datasets, implying that no aggregated global network is used. When $\lambda \in (0, 1)$, the performances are consistently better than FedBYOL ($\lambda = 0$) as both local and global knowledge are effectively aggregated. These analyses are mainly suitable for our experiment setting. The range values of $\lambda$ depend on the characteristics of data and the hyper-parameters (e.g., local epoch) of FL settings. A practical way to tune $\lambda$ manually is to understand the divergence by running the algorithm for several rounds and choose the $\lambda$ that scales $\mu$ to (0.5, 1). Nevertheless, we recommend using autoscaler and provide ablation study of $\tau$ of autoscaler in Figure \ref{fig:autoscaler} in Appendix \ref{apx:experimental-results}.

\textbf{Constant Values of $\mu$} \, We further demonstrate the necessity of dynamic EMA by comparing with using constant values of $\mu$ in Eqn \ref{eq:encoder} and \ref{eq:predictor}. Figure \ref{fig:const-ema} shows that a good choice of constant $\mu$ can outperform FedBYOL, but FedEMA outperforms using constant $\mu$ for the online encoder, predictor, or applying both. We also provide results that encoder and predictor use different $\mu$ in Appendix \ref{apx:experimental-results}. 

% Figure \ref{fig:const-ema} using either constant $\mu$ for only online encoder, only predictor, or both  has lower performance compared to our method. Besides, our method also outperforms applying both constant $\mu_{o}$ and $\mu_{p}$ in Figure \ref{fig:ema-encoder-predictor}.

\textbf{Non-IID Level} \, Figure \ref{fig:non-iid-level} compares the performance of different non-IID levels, ranging from 2 to 10 classes per client on the CIFAR-10 dataset. We use autoscaler for these experiments. FedEMA consistently outperforms FedBYOL in these settings. 
% As expected that the performances increases with the decrease of non-IID levels and the 
% $\lambda = 1$ for 2 and 4 classes per client and $\lambda = 0.8$ for 6, 8, and 10 classes per client

% FedEMA consistently outperforms FedBYOL in these settings. Moreover, the performances of these methods increase as the non-IID level decreases.

% We provide more results of varying $E$ in Figure \ref{fig:non-iid-local-epoch} in the Appendix.

% Table \ref{fig:comp_vs_comm} compares rounds to reach a target accuracy with different local epochs $E$ using FedEMA. Increasing $E$ reduces communication cost as the needed communication rounds decrease, but it requires a higher computation cost. For example, compared with $E = 5$ that needs 85 rounds to reach 80\% with 425 epochs of computation, $E = 20$ only uses 40 rounds but needs 800 epochs computation cost. These results indicate the trade-off between communication cost and computation cost. We provide more results of varying $E$ in the Appendix.

\section{Conclusion}

We uncover important insights of federated self-supervised learning (FedSSL) from in-depth empirical studies, using a newly introduced generalized FedSSL framework. Inspired by the insights, we propose a new method, Federated Divergence-aware Exponential Moving Average update (FedEMA), to further address the non-IID data challenge. Our experiments and ablations demonstrate that FedEMA outperforms existing methods in a wide range of settings. In the future, we plan to implement FedSSL and FedEMA on larger-scale datasets. We hope that this study will provide useful insights for future research.

% we present a new federated self-supervised learning framework and conduct empirical studies to generate unique insights on FedSSL. Based on these insights, we propose a FedEMA to update local models of clients adaptively with the dynamic EMA of the global model. Extensive experiments and ablations demonstrate the significance of FedEMA. We believe that the evidence of this study will provide useful insights for future research.

\section{Reproducibility Statement}

To facilitate reproducibility of experiment results, we first provide basic experimental setups in Section \ref{sec:experimental-setup}, including datasets, implementation details, and evaluation protocols. Then, we describe more experimental details in Appendix \ref{apx:experimental-details}, including datasets, data transformation, network architecture, training details, and default settings. Also, we indicate the settings and hyper-parameters of experiments when their settings are different from the default. Moreover, we plan to open-source the codes in the future.

\subsubsection*{Acknowledgments}
We would like to thank reviewers of ICLR 2022 for their constructive and helpful feedback. This study is in part supported by the RIE2020 Industry Alignment Fund – Industry Collaboration Projects (IAF-ICP) Funding Initiative, as well as cash and in-kind contribution from the industry partner(s); the National Research Foundation, Singapore under its Energy Programme (EP Award NRF2017EWT-EP003-023) administrated by the Energy Market Authority of Singapore, and its Energy Research Test-Bed and Industry Partnership Funding Initiative, part of the Energy Grid (EG) 2.0 programme, and its Central Gap Fund (“Central Gap” Award No. NRF2020NRF-CG001-027); Singapore MOE under its Tier 1 grant call, Reference number RG96/20.

\bibliography{iclr2022_conference}

\begin{thebibliography}{46}
\providecommand{\natexlab}[1]{#1}
\providecommand{\url}[1]{\texttt{#1}}
\expandafter\ifx\csname urlstyle\endcsname\relax
  \providecommand{\doi}[1]{doi: #1}\else
  \providecommand{\doi}{doi: \begingroup \urlstyle{rm}\Url}\fi

\bibitem[Bachman et~al.(2019)Bachman, Hjelm, and
  Buchwalter]{bachman2019contrastive}
Philip Bachman, R~Devon Hjelm, and William Buchwalter.
\newblock Learning representations by maximizing mutual information across
  views.
\newblock In H.~Wallach, H.~Larochelle, A.~Beygelzimer, F.~d\textquotesingle
  Alch\'{e}-Buc, E.~Fox, and R.~Garnett (eds.), \emph{Advances in Neural
  Information Processing Systems}, volume~32. Curran Associates, Inc., 2019.
\newblock URL
  \url{https://proceedings.neurips.cc/paper/2019/file/ddf354219aac374f1d40b7e760ee5bb7-Paper.pdf}.

\bibitem[Caldas et~al.(2018)Caldas, Duddu, Wu, Li, Kone{\v{c}}n{\`y}, McMahan,
  Smith, and Talwalkar]{caldas2018leaf}
Sebastian Caldas, Sai Meher~Karthik Duddu, Peter Wu, Tian Li, Jakub
  Kone{\v{c}}n{\`y}, H~Brendan McMahan, Virginia Smith, and Ameet Talwalkar.
\newblock Leaf: A benchmark for federated settings.
\newblock \emph{arXiv preprint arXiv:1812.01097}, 2018.

\bibitem[Chen et~al.(2020{\natexlab{a}})Chen, Kornblith, Norouzi, and
  Hinton]{chen2020simclr}
Ting Chen, Simon Kornblith, Mohammad Norouzi, and Geoffrey Hinton.
\newblock A simple framework for contrastive learning of visual
  representations.
\newblock In \emph{International conference on machine learning}, pp.\
  1597--1607. PMLR, 2020{\natexlab{a}}.

\bibitem[Chen \& He(2021)Chen and He]{chen2020simsiam}
Xinlei Chen and Kaiming He.
\newblock Exploring simple siamese representation learning.
\newblock In \emph{Proceedings of the IEEE/CVF Conference on Computer Vision
  and Pattern Recognition (CVPR)}, pp.\  15750--15758, June 2021.

\bibitem[Chen et~al.(2020{\natexlab{b}})Chen, Fan, Girshick, and
  He]{chen2020mocov2}
Xinlei Chen, Haoqi Fan, Ross Girshick, and Kaiming He.
\newblock Improved baselines with momentum contrastive learning.
\newblock \emph{arXiv preprint arXiv:2003.04297}, 2020{\natexlab{b}}.

\bibitem[Custers et~al.(2019)Custers, Sears, Dechesne, Georgieva, Tani, and
  van~der Hof]{gdpr}
Bart Custers, Alan~M. Sears, Francien Dechesne, Ilina Georgieva, Tommaso Tani,
  and Simone van~der Hof.
\newblock \emph{EU Personal Data Protection in Policy and Practice}.
\newblock Springer, 2019.

\bibitem[Doersch et~al.(2015)Doersch, Gupta, and
  Efros]{doersch2015unsup-context-prediction}
Carl Doersch, Abhinav Gupta, and Alexei~A Efros.
\newblock Unsupervised visual representation learning by context prediction.
\newblock In \emph{Proceedings of the IEEE international conference on computer
  vision}, pp.\  1422--1430, 2015.

\bibitem[Gidaris et~al.(2018)Gidaris, Singh, and
  Komodakis]{gidaris2018unsup-image-rotation}
Spyros Gidaris, Praveer Singh, and Nikos Komodakis.
\newblock Unsupervised representation learning by predicting image rotations.
\newblock \emph{arXiv preprint arXiv:1803.07728}, 2018.

\bibitem[Grill et~al.(2020)Grill, Strub, Altch\'{e}, Tallec, Richemond,
  Buchatskaya, Doersch, Avila~Pires, Guo, Gheshlaghi~Azar, Piot, kavukcuoglu,
  Munos, and Valko]{grill2020byol}
Jean-Bastien Grill, Florian Strub, Florent Altch\'{e}, Corentin Tallec, Pierre
  Richemond, Elena Buchatskaya, Carl Doersch, Bernardo Avila~Pires, Zhaohan
  Guo, Mohammad Gheshlaghi~Azar, Bilal Piot, koray kavukcuoglu, Remi Munos, and
  Michal Valko.
\newblock Bootstrap your own latent - a new approach to self-supervised
  learning.
\newblock In H.~Larochelle, M.~Ranzato, R.~Hadsell, M.~F. Balcan, and H.~Lin
  (eds.), \emph{Advances in Neural Information Processing Systems}, volume~33,
  pp.\  21271--21284. Curran Associates, Inc., 2020.
\newblock URL
  \url{https://proceedings.neurips.cc/paper/2020/file/f3ada80d5c4ee70142b17b8192b2958e-Paper.pdf}.

\bibitem[Hanzely et~al.(2020)Hanzely, Hanzely, Horv{\'a}th, and
  Richt{\'a}rik]{hanzely2020lower}
Filip Hanzely, Slavom{\'\i}r Hanzely, Samuel Horv{\'a}th, and Peter
  Richt{\'a}rik.
\newblock Lower bounds and optimal algorithms for personalized federated
  learning.
\newblock In \emph{Advances in Neural Information Processing Systems}. Curran
  Associates, Inc., 2020.

\bibitem[He et~al.(2016)He, Zhang, Ren, and Sun]{he2016resnet}
Kaiming He, Xiangyu Zhang, Shaoqing Ren, and Jian Sun.
\newblock Deep residual learning for image recognition.
\newblock In \emph{Proceedings of the IEEE conference on computer vision and
  pattern recognition}, pp.\  770--778, 2016.

\bibitem[He et~al.(2020)He, Fan, Wu, Xie, and Girshick]{he2020moco}
Kaiming He, Haoqi Fan, Yuxin Wu, Saining Xie, and Ross Girshick.
\newblock Momentum contrast for unsupervised visual representation learning.
\newblock In \emph{Proceedings of the IEEE/CVF Conference on Computer Vision
  and Pattern Recognition}, pp.\  9729--9738, 2020.

\bibitem[Jeong et~al.(2021)Jeong, Yoon, Yang, and Hwang]{jeong2021fedsemi}
Wonyong Jeong, Jaehong Yoon, Eunho Yang, and Sung~Ju Hwang.
\newblock Federated semi-supervised learning with inter-client consistency \&
  disjoint learning.
\newblock In \emph{International Conference on Learning Representations}, 2021.
\newblock URL \url{https://openreview.net/forum?id=ce6CFXBh30h}.

\bibitem[Jin et~al.(2020{\natexlab{a}})Jin, Wei, Liu, and
  Yang]{jin2020fed-unlabeled-survey}
Yilun Jin, Xiguang Wei, Yang Liu, and Qiang Yang.
\newblock Towards utilizing unlabeled data in federated learning: A survey and
  prospective.
\newblock \emph{arXiv e-prints}, pp.\  arXiv--2002, 2020{\natexlab{a}}.

\bibitem[Jin et~al.(2020{\natexlab{b}})Jin, Wei, Liu, and
  Yang]{jin2020survey-fedsemi}
Yilun Jin, Xiguang Wei, Yang Liu, and Qiang Yang.
\newblock A survey towards federated semi-supervised learning.
\newblock \emph{arXiv preprint arXiv:2002.11545}, 2020{\natexlab{b}}.

\bibitem[Kairouz et~al.(2019)Kairouz, McMahan, Avent, Bellet, Bennis, Bhagoji,
  Bonawitz, Charles, Cormode, Cummings, et~al.]{kairouz2019advances}
Peter Kairouz, H~Brendan McMahan, Brendan Avent, Aur{\'e}lien Bellet, Mehdi
  Bennis, Arjun~Nitin Bhagoji, Kallista Bonawitz, Zachary Charles, Graham
  Cormode, Rachel Cummings, et~al.
\newblock Advances and open problems in federated learning.
\newblock \emph{arXiv preprint arXiv:1912.04977}, 2019.

\bibitem[Kolesnikov et~al.(2019)Kolesnikov, Zhai, and
  Beyer]{kolesnikov2019revisiting}
Alexander Kolesnikov, Xiaohua Zhai, and Lucas Beyer.
\newblock Revisiting self-supervised visual representation learning.
\newblock In \emph{Proceedings of the IEEE/CVF Conference on Computer Vision
  and Pattern Recognition}, pp.\  1920--1929, 2019.

\bibitem[Krizhevsky et~al.(2009)Krizhevsky, Hinton, et~al.]{cifar10-2009}
Alex Krizhevsky, Geoffrey Hinton, et~al.
\newblock Learning multiple layers of features from tiny images.
\newblock 2009.

\bibitem[Li et~al.(2020{\natexlab{a}})Li, Sahu, Talwalkar, and
  Smith]{Li2020FedChallenges}
Tian Li, Anit~Kumar Sahu, Ameet Talwalkar, and Virginia Smith.
\newblock Federated learning: Challenges, methods, and future directions.
\newblock \emph{IEEE Signal Processing Magazine}, 37:\penalty0 50--60,
  2020{\natexlab{a}}.

\bibitem[Li et~al.(2020{\natexlab{b}})Li, Sahu, Zaheer, Sanjabi, Talwalkar, and
  Smith]{fedprox}
Tian Li, Anit~Kumar Sahu, Manzil Zaheer, Maziar Sanjabi, Ameet Talwalkar, and
  Virginia Smith.
\newblock Federated optimization in heterogeneous networks.
\newblock \emph{Proceedings of Machine Learning and Systems}, 2:\penalty0
  429--450, 2020{\natexlab{b}}.

\bibitem[Loshchilov \& Hutter(2017)Loshchilov and Hutter]{sgdr-cosine-lr}
Ilya Loshchilov and Frank Hutter.
\newblock {SGDR:} stochastic gradient descent with warm restarts.
\newblock In \emph{5th International Conference on Learning Representations,
  {ICLR} 2017, Toulon, France, April 24-26, 2017, Conference Track
  Proceedings}, 2017.
\newblock URL \url{https://openreview.net/forum?id=Skq89Scxx}.

\bibitem[Luo et~al.(2019)Luo, Wu, Luo, Huang, Huang, Liu, and
  Yang]{luo2019real-obj-detect}
Jiahuan Luo, Xueyang Wu, Yun Luo, Anbu Huang, Yunfeng Huang, Yang Liu, and
  Qiang Yang.
\newblock Real-world image datasets for federated learning.
\newblock \emph{arXiv preprint arXiv:1910.11089}, 2019.

\bibitem[Mansour et~al.(2020)Mansour, Mohri, Ro, and Suresh]{mansour2020three}
Yishay Mansour, Mehryar Mohri, Jae Ro, and Ananda~Theertha Suresh.
\newblock Three approaches for personalization with applications to federated
  learning.
\newblock \emph{arXiv preprint arXiv:2002.10619}, 2020.

\bibitem[McMahan et~al.(2017)McMahan, Moore, Ramage, Hampson, and
  y~Arcas]{fedavg}
Brendan McMahan, Eider Moore, Daniel Ramage, Seth Hampson, and
  Blaise~Ag{\"{u}}era y~Arcas.
\newblock Communication-efficient learning of deep networks from decentralized
  data.
\newblock In Aarti Singh and Xiaojin~(Jerry) Zhu (eds.), \emph{Proceedings of
  the 20th International Conference on Artificial Intelligence and Statistics,
  {AISTATS} 2017, 20-22 April 2017, Fort Lauderdale, FL, {USA}}, volume~54 of
  \emph{Proceedings of Machine Learning Research}, pp.\  1273--1282. {PMLR},
  2017.
\newblock URL \url{http://proceedings.mlr.press/v54/mcmahan17a.html}.

\bibitem[Noroozi \& Favaro(2016)Noroozi and Favaro]{noroozi2016jigsaw}
Mehdi Noroozi and Paolo Favaro.
\newblock Unsupervised learning of visual representations by solving jigsaw
  puzzles.
\newblock In \emph{European conference on computer vision}, pp.\  69--84.
  Springer, 2016.

\bibitem[Oord et~al.(2018)Oord, Li, and
  Vinyals]{oord2018contrastive-predictive-coding}
Aaron van~den Oord, Yazhe Li, and Oriol Vinyals.
\newblock Representation learning with contrastive predictive coding.
\newblock \emph{arXiv preprint arXiv:1807.03748}, 2018.

\bibitem[Paszke et~al.(2017)Paszke, Gross, Chintala, Chanan, Yang, DeVito, Lin,
  Desmaison, Antiga, and Lerer]{paszke2017pytorch}
Adam Paszke, Sam Gross, Soumith Chintala, Gregory Chanan, Edward Yang, Zachary
  DeVito, Zeming Lin, Alban Desmaison, Luca Antiga, and Adam Lerer.
\newblock Automatic differentiation in pytorch.
\newblock 2017.

\bibitem[Pathak et~al.(2016)Pathak, Krahenbuhl, Donahue, Darrell, and
  Efros]{pathak2016inpainting}
Deepak Pathak, Philipp Krahenbuhl, Jeff Donahue, Trevor Darrell, and Alexei~A
  Efros.
\newblock Context encoders: Feature learning by inpainting.
\newblock In \emph{Proceedings of the IEEE conference on computer vision and
  pattern recognition}, pp.\  2536--2544, 2016.

\bibitem[Peng et~al.(2020)Peng, Huang, Zhu, and Saenko]{peng2020fuda}
Xingchao Peng, Zijun Huang, Yizhe Zhu, and Kate Saenko.
\newblock Federated adversarial domain adaptation.
\newblock In \emph{International Conference on Learning Representations}, 2020.
\newblock URL \url{https://openreview.net/forum?id=HJezF3VYPB}.

\bibitem[Tan et~al.(2021)Tan, Yu, Cui, and Yang]{tan2021personlized-fl-survey}
Alysa~Ziying Tan, Han Yu, Lizhen Cui, and Qiang Yang.
\newblock Towards personalized federated learning.
\newblock \emph{arXiv preprint arXiv:2103.00710}, 2021.

\bibitem[Tian et~al.(2021)Tian, Chen, and Ganguli]{tian21directpred}
Yuandong Tian, Xinlei Chen, and Surya Ganguli.
\newblock Understanding self-supervised learning dynamics without contrastive
  pairs.
\newblock In Marina Meila and Tong Zhang (eds.), \emph{Proceedings of the 38th
  International Conference on Machine Learning}, volume 139 of
  \emph{Proceedings of Machine Learning Research}, pp.\  10268--10278. PMLR,
  18--24 Jul 2021.
\newblock URL \url{https://proceedings.mlr.press/v139/tian21a.html}.

\bibitem[van Berlo et~al.(2020)van Berlo, Saeed, and Ozcelebi]{van2020fedae}
Bram van Berlo, Aaqib Saeed, and Tanir Ozcelebi.
\newblock Towards federated unsupervised representation learning.
\newblock In \emph{Proceedings of the Third ACM International Workshop on Edge
  Systems, Analytics and Networking}, pp.\  31--36, 2020.

\bibitem[Wang et~al.(2020)Wang, Yurochkin, Sun, Papailiopoulos, and
  Khazaeni]{Wang2020fedma}
Hongyi Wang, Mikhail Yurochkin, Yuekai Sun, Dimitris Papailiopoulos, and
  Yasaman Khazaeni.
\newblock Federated learning with matched averaging.
\newblock In \emph{International Conference on Learning Representations}, 2020.
\newblock URL \url{https://openreview.net/forum?id=BkluqlSFDS}.

\bibitem[Wu et~al.(2018)Wu, Xiong, Yu, and Lin]{wu2018unsupervised-instance}
Zhirong Wu, Yuanjun Xiong, Stella~X Yu, and Dahua Lin.
\newblock Unsupervised feature learning via non-parametric instance
  discrimination.
\newblock In \emph{Proceedings of the IEEE Conference on Computer Vision and
  Pattern Recognition}, pp.\  3733--3742, 2018.

\bibitem[Yan et~al.(2020)Yan, Acuna, and Fidler]{yan2020neural}
Xi~Yan, David Acuna, and Sanja Fidler.
\newblock Neural data server: A large-scale search engine for transfer learning
  data.
\newblock In \emph{Proceedings of the IEEE/CVF Conference on Computer Vision
  and Pattern Recognition}, pp.\  3893--3902, 2020.

\bibitem[yuyang deng et~al.(2021)yuyang deng, Kamani, and
  Mahdavi]{deng2021adaptive}
yuyang deng, Mohammad~Mahdi Kamani, and Mehrdad Mahdavi.
\newblock Adaptive personalized federated learning, 2021.
\newblock URL \url{https://openreview.net/forum?id=g0a-XYjpQ7r}.

\bibitem[Zhai et~al.(2019)Zhai, Oliver, Kolesnikov, and Beyer]{zhai2019s4l}
Xiaohua Zhai, Avital Oliver, Alexander Kolesnikov, and Lucas Beyer.
\newblock S4l: Self-supervised semi-supervised learning.
\newblock In \emph{Proceedings of the IEEE/CVF International Conference on
  Computer Vision}, pp.\  1476--1485, 2019.

\bibitem[Zhang et~al.(2020{\natexlab{a}})Zhang, Kuang, You, Shen, Xiao, Zhang,
  Wu, Zhuang, and Li]{zhang2020fedca}
Fengda Zhang, Kun Kuang, Zhaoyang You, Tao Shen, Jun Xiao, Yin Zhang, Chao Wu,
  Yueting Zhuang, and Xiaolin Li.
\newblock Federated unsupervised representation learning.
\newblock \emph{arXiv preprint arXiv:2010.08982}, 2020{\natexlab{a}}.

\bibitem[Zhang et~al.(2016)Zhang, Isola, and Efros]{zhang2016colorful-image}
Richard Zhang, Phillip Isola, and Alexei~A Efros.
\newblock Colorful image colorization.
\newblock In \emph{European conference on computer vision}, pp.\  649--666.
  Springer, 2016.

\bibitem[Zhang et~al.(2020{\natexlab{b}})Zhang, Yao, Yang, Yan, Gonzalez, and
  Mahoney]{zhang2020benchmark-fedsemi}
Zhengming Zhang, Zhewei Yao, Yaoqing Yang, Yujun Yan, Joseph~E Gonzalez, and
  Michael~W Mahoney.
\newblock Benchmarking semi-supervised federated learning.
\newblock \emph{arXiv preprint arXiv:2008.11364}, 17, 2020{\natexlab{b}}.

\bibitem[Zhao et~al.(2018)Zhao, Li, Lai, Suda, Civin, and
  Chandra]{zhao2018non-iid}
Yue Zhao, Meng Li, Liangzhen Lai, Naveen Suda, Damon Civin, and Vikas Chandra.
\newblock Federated learning with non-iid data.
\newblock \emph{CoRR}, abs/1806.00582, 2018.
\newblock URL \url{http://arxiv.org/abs/1806.00582}.

\bibitem[Zhuang et~al.(2020)Zhuang, Wen, Zhang, Gan, Yin, Zhou, Zhang, and
  Yi]{zhuang2020fedreid}
Weiming Zhuang, Yonggang Wen, Xuesen Zhang, Xin Gan, Daiying Yin, Dongzhan
  Zhou, Shuai Zhang, and Shuai Yi.
\newblock Performance optimization of federated person re-identification via
  benchmark analysis.
\newblock In \emph{Proceedings of the 28th ACM International Conference on
  Multimedia}, pp.\  955--963, 2020.

\bibitem[Zhuang et~al.(2021{\natexlab{a}})Zhuang, Gan, Wen, Zhang, and
  Yi]{zhuang2021fedu}
Weiming Zhuang, Xin Gan, Yonggang Wen, Shuai Zhang, and Shuai Yi.
\newblock Collaborative unsupervised visual representation learning from
  decentralized data.
\newblock In \emph{Proceedings of the IEEE/CVF International Conference on
  Computer Vision}, pp.\  4912--4921, 2021{\natexlab{a}}.

\bibitem[Zhuang et~al.(2021{\natexlab{b}})Zhuang, Gan, Wen, Zhang, Zhang, and
  Yi]{zhuang2021fedfr}
Weiming Zhuang, Xin Gan, Yonggang Wen, Xuesen Zhang, Shuai Zhang, and Shuai Yi.
\newblock Towards unsupervised domain adaptation for deep face recognition
  under privacy constraints via federated learning.
\newblock \emph{arXiv preprint arXiv:2105.07606}, 2021{\natexlab{b}}.

\bibitem[Zhuang et~al.(2021{\natexlab{c}})Zhuang, Wen, and
  Zhang]{zhuang2021fedureid}
Weiming Zhuang, Yonggang Wen, and Shuai Zhang.
\newblock Joint optimization in edge-cloud continuum for federated unsupervised
  person re-identification.
\newblock In \emph{Proceedings of the 29th ACM International Conference on
  Multimedia}, pp.\  433--441, 2021{\natexlab{c}}.

\bibitem[Zhuang et~al.(2022)Zhuang, Gan, Wen, and Zhang]{zhuang2022easyfl}
Weiming Zhuang, Xin Gan, Yonggang Wen, and Shuai Zhang.
\newblock Easyfl: A low-code federated learning platform for dummies.
\newblock \emph{IEEE Internet of Things Journal}, 2022.

\end{thebibliography}
\bibliographystyle{iclr2022_conference}

\clearpage

\appendix

\section{Differences of Self-supervised Learning Methods}
\label{apx:ssl-methods}

\begin{figure}[h]
   \begin{center}
   \includegraphics[width=0.8\linewidth]{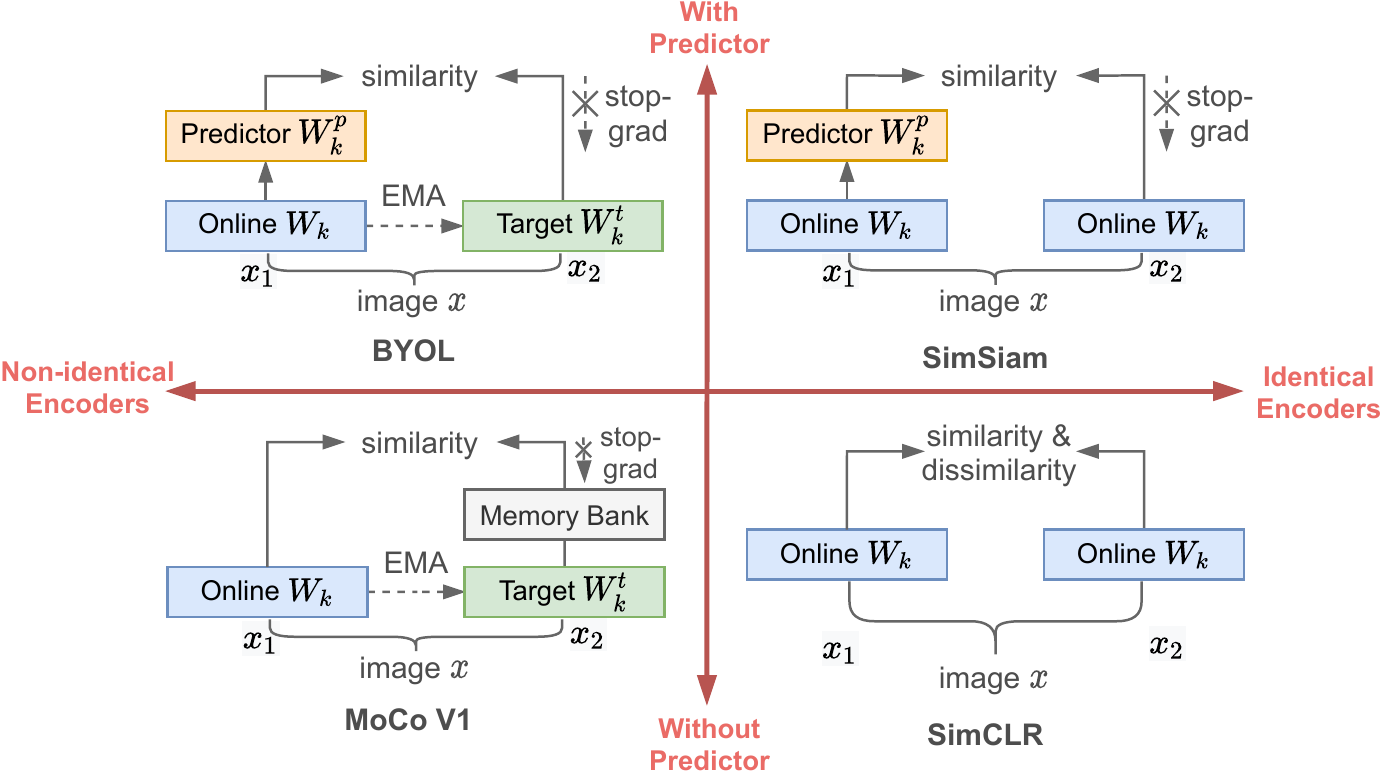}
      \caption{Illustration of differences among four Self-supervised Learning (SSL) methods.}
   \label{fig:ssl-methods}
   \end{center}
\end{figure}

We study four SSL methods using the FedSSL framework in Section \ref{sec:empirical-study}. These four SSL methods have two major differences that impact the executions of local training, model communication, and model aggregation. Figure \ref{fig:ssl-methods} depicts these differences: 1) BYOL and SimSiam have predictors, whereas MoCo and SimCLR do not have them; 2) SimSiam and SimCLR share weights between two encoders, whereas BYOL and MoCo have different parameters for the online and target encoders. 

\section{Experimental Details}
\label{apx:experimental-details}

In this section, we provide more details about the dataset, network architecture, and training and evaluation setups. 

\subsection{Data}

\textbf{Datasets} \, CIFAR-10 and CIFAR-100 are two popular image datasets \citep{cifar10-2009}. Both datasets consist of 50,000 training images and 10,000 testing images. CIFAR-10 contains 10 classes, where each class has 5,000 training images and 1,000 testing images. While CIFAR-100 contains 100 classes, where each class has 500 training images and 100 testing images. To simulate federated learning, we equally split the training set into $K$ clients. We simulate non-IID data using label heterogeneity --- data among clients is more skewed when each client contains less number of classes. Hence, we simulate different levels of non-IID data with $l$ number of classes per client, where $l = \{2, 4, 6, 8, 10\}$ for CIFAR-10 and $l = \{20, 40, 60, 80, 100\}$ for CIFAR-100. For example, when simulating 5 clients with $l = 4$ classes per client in CIFAR-10, we need $5 \times 4 = 20$ total sets of data over 10 classes. Thus, we split the training images of each class equally into two sets (2,500 images in each set) and assign random four sets without overlapping classes to a client. The setting is IID when each client contains all classes of a dataset. By default, we run experiments with $K = 5$ clients with non-IID setting $l = 2$ classes per client for CIFAR-10 dataset and $l = 20$ classes per client for CIFAR-100 dataset.

\textbf{Transformation} \, In local training of the FedSSL framework, we take two augmentations of each image as the inputs for online and target networks, respectively. We obtain the augmentations by transforming the images with a set of transformations: For SimCLR, BYOL, SimSiam, and MoCoV2, we adopt the transformations from \citet{chen2020simclr}; For MoCoV1, we use the transformation described in its paper \citep{he2020moco}.

\begingroup
\setlength{\tabcolsep}{0.4em}
\begin{table*}[t]
   \caption{Top-1 accuracy comparison under linear evaluation protocol on CIFAR datasets. Our proposed FedEMA outperforms all other methods on non-IID settings.}
   \label{tab:linear-eval-full}
   \begin{center}
   \begin{tabular}{llccgccg}
   \hline
   \multicolumn{1}{l}{\multirow{2}{*}{Method}} &
   \multicolumn{1}{l}{\multirow{2}{*}{Architecture}} &
   \multicolumn{1}{c}{\multirow{2}{*}{Param.}} &
   \multicolumn{2}{c}{CIFAR-10 (\%)} &
   \multicolumn{1}{c}{} &
   \multicolumn{2}{c}{CIFAR-100 (\%)}
   \\ 
   \cline{4-5} \\[-4.1mm] \cline{7-8} \\[-4.1mm]
   \multicolumn{1}{c}{} & & & IID & \cellcolor{white} Non-IID & & IID &  \cellcolor{white} Non-IID
   \\
   % \hline 
   \Xhline{3\arrayrulewidth}
   Standalone training & ResNet-18 & 11M & 82.42 & 74.95 & & 53.88 & 52.37 \\
   FedSimCLR & ResNet-18 & 11M & 82.15 & 78.09 & & 56.39 & 55.58 \\
   FedMoCoV1 & ResNet-18 & 11M & 83.63 & 78.21 & & \textbf{59.58} & 56.98 \\
   FedMoCoV2 & ResNet-18 & 11M & 84.25 & 79.14 & & 58.71 & 57.47 \\
   FedSimSiam & ResNet-18 & 11M & 81.46 & 76.27 & & 49.92 & 48.94 \\
   FedBYOL & ResNet-18 & 11M & 84.29 & 79.44 & & 54.24 & 57.51 \\
   FedU \citep{zhuang2021fedu} & ResNet-18 & 11M & 83.96 & 80.52 & & 54.82 & 57.21 \\
   FedEMA predictor only (ours) & ResNet-18 & 11M & 84.97 & 81.13 & & 55.52 & 57.53 \\
   FedEMA encoder only (ours) & ResNet-18 & 11M & 82.88 & 82.39 & & 56.06 & 59.74 \\
   FedEMA ($\lambda = 0.8$) & ResNet-18 & 11M & \textbf{85.59} & \textbf{82.77} & & 57.86 & \textbf{61.21} \\
   % FedEMA (autoscaler $\tau = 0.7$) & ResNet-18 & 11M & 86.17 & 83.10 & & 58.31 & 61.87 \\
   FedEMA (autoscaler, $\tau = 0.7$) & ResNet-18 & 11M & \textbf{86.26} & \textbf{83.34} & & 58.55 & \textbf{61.78} \\
   % FedEMA (autoscaler, $\tau = 0.8$) & ResNet-18 & 11M & 85.58 & 83.36 & & 57.59 & 60.91 \\

   % FedEMA (autoscaler) & ResNet-18 & 11M & \textbf{85.58} & \textbf{83.36} & & 57.59 & \textbf{60.91} \\
   % FedU \citep{zhuang2021fedu} & ResNet-18 & 11M & 85.21 & 78.71 & & & 56.52 & 57.08 \\ 
   \hline
   Standalone training & ResNet-50 & 23M & 83.16 & 77.84 & & 57.21 & 55.16 \\
   FedSimCLR & ResNet-50 & 23M & 82.24 & 80.37 & & 57.46 & 56.88 \\
   FedMoCoV1 & ResNet-50 & 23M & 87.19 & 82.18 & & \textbf{64.74} & 59.73 \\
   FedMoCoV2 & ResNet-50 & 23M & \textbf{87.19} & 79.62 & & 63.75 & 59.52 \\
   FedSimSiam & ResNet-50 & 23M & 79.64 & 76.7 & & 46.28 & 48.8 \\
   FedBYOL & ResNet-50 & 23M & 83.90 & 81.33 & & 57.75 & 59.53 \\
   FedCA \citep{zhang2020fedca}  & ResNet-50 & 23M & 71.25 & 68.01 & & 43.30 & 42.34 \\ 
   FedU \citep{zhuang2021fedu} & ResNet-50 & 23M & 86.48 & 83.25 & & 59.51 & 61.94 \\
   FedEMA predictor only (ours) & ResNet-50 & 23M & 83.66 & 81.78 & & 57.79 & 60.11 \\
   FedEMA encoder only (ours) & ResNet-50 & 23M & 84.66 & 84.91 & & 58.52 & 62.51 \\
   FedEMA ($\lambda = 0.8$) & ResNet-50 & 23M & 86.12 & \textbf{85.29} & & 60.96 & \textbf{62.53} \\ 
   
   FedEMA (autoscaler, $\tau = 0.7$) & ResNet-50 & 23M & 85.08 & \textbf{84.31} & & 59.48 & \textbf{62.77} \\
   % FedEMA (autoscaler, $\tau = 0.8$) & ResNet-50 & 23M & 85.29 & 82.86 & & 58.86 & 61.70 \\

   % FedEMA (autoscaler) & ResNet-50 & 23M & 85.29 & 82.86 & & 58.86 & 61.70 \\ 

   %  &  % \hline
   % \multicolumn{8}{l}{\textit{Upper-bound methods: centralized unsupervised learning and supervised federated learning}} \\
   \hline
   % FedAvg (Supervised) & ResNet-18 & 11M & 91.97 & 74.21 & & 65.76 & 64.06 \\
   BYOL (Centralized) & ResNet-18 & 11M & 90.46 & \cellcolor{white}  - & & 65.54 & \cellcolor{white} - \\
   % FedAvg (Supervised) & ResNet-50 & 23M & 91.51 & 67.74 & & 65.77 & 64.38 \\
   BYOL (Centralized) & ResNet-50 & 23M & 91.85 & \cellcolor{white} - & & 66.51 & \cellcolor{white} - \\
   \hline
   \end{tabular}
   \end{center}
\end{table*}    
\endgroup

\begingroup
\setlength{\tabcolsep}{0.4em}
\begin{table*}[t]
   \begin{center}
   \begin{tabular}{lccccccccccc}
   \hline
   \multicolumn{1}{l}{\multirow{3}{*}{Method}} &
   \multicolumn{5}{c}{5/20 clients (\%)} &
   \multicolumn{1}{c}{} &
   \multicolumn{5}{c}{8/80 clients (\%)}
   \\ 
   \cline{2-6} \cline{8-12}
   \multicolumn{1}{c}{} & 
   \multicolumn{2}{c}{CIFAR-10} &
   \multicolumn{1}{c}{} &
   \multicolumn{2}{c}{CIFAR-100} & &
   \multicolumn{2}{c}{CIFAR-10} &
   \multicolumn{1}{c}{} &
   \multicolumn{2}{c}{CIFAR-100}
   \\
   \cline{2-3} \cline{5-6} \cline{8-9} \cline{11-12}
   \multicolumn{1}{c}{} & IID & Non-IID & & IID & Non-IID & & IID &Non-IID & & IID & Non-IID
   \\
   % \hline 
   \Xhline{3\arrayrulewidth}
   FedBYOL & 83.25 & 74.92 & & 49.49 & 47.09 & & 73.58 & 63.28 & & 41.19 & 41.58 \\
   FedEMA (ours) & \textbf{84.98} & \textbf{75.77} & & \textbf{55.41} & \textbf{52.78} & & \textbf{73.96} & \textbf{64.19} & & \textbf{41.97} & \textbf{43.05} \\
   % FedBYOL & 84.29 & 79.44 & & 54.24 & 57.51 \\ 
   % FedEMA & \textbf{0.7577} & \textbf{83.36} & & \textbf{57.29} & \textbf{60.91} \\ 
   \hline
   \end{tabular}
   \caption{Top-1 accuracy comparison on larger numbers of clients with client subsampling: 1) randomly selecting 5 out of 20 clients per round (5/20); 2) randomly selecting 8 out of 80 clients per round (8/80). FedEMA, trained with autoscaler, consistently outperforms FedBYOL in both settings.}
   \label{tab:scalability}
   \end{center}
\end{table*}    
\endgroup

% (0.8377 + 0.8354) / 2 = 0.8366
% (0.8103 + 0.8252) / 2 = 0.8178
% (0.5836 + 0.5785 + 0.5716) / 3 = 0.5779
% (0.5958 + 0.6036 + 0.6040) / 3 = 0.6011

\begin{table*}[t]
   \caption{Top-1 accuracy comparison on using 1\% and 10\% of labeled data for semi-supervised learning on the non-IID settings of CIFAR datasets. FedEMA outperforms all other methods.}
   \label{tab:semi-sup-full}
   \begin{center}
   \begin{tabular}{llccccccccc}
   \hline
   \multicolumn{1}{l}{\multirow{2}{*}{Method}} &
   \multicolumn{1}{l}{\multirow{2}{*}{Architecture}} &
   \multicolumn{1}{c}{\multirow{2}{*}{Param.}} &
   \multicolumn{2}{c}{CIFAR-10 (\%)} &
   \multicolumn{1}{c}{} &
   \multicolumn{2}{c}{CIFAR-100 (\%)}
   \\ 
   \cline{4-5} \cline{7-8}
   \multicolumn{1}{c}{} & & & 1\% & 10\% & & 1\% & 10\%
   \\
   \Xhline{3\arrayrulewidth}

   Standalone training & ResNet-18 & 11M & 61.37 & 69.06 & & 21.37 & 39.99 \\
   FedSimCLR & ResNet-18 & 11M & 63.79 & 73.49 & & 21.55 & 41.90 \\
   FedMoCoV1 & ResNet-18 & 11M & 60.57 & 73.95 & & 21.83 & 43.49 \\
   FedMoCoV2 & ResNet-18 & 11M & 62.89 & 73.65 & & 26.93 & 45.27 \\
   FedSimSiam & ResNet-18 & 11M & 67.57 & 74.96 & & 25.13 & 41.96 \\
   FedBYOL & ResNet-18 & 11M & 70.48 & 76.95 & & 30.21 & 47.07 \\
   FedU \citep{zhuang2021fedu} & ResNet-18 & 11M & 69.52 & 77.06 & & 29.00 & 46.67 \\
   FedEMA ($\lambda = 1$) & ResNet-18 & 11M & \textbf{72.78} & \textbf{79.01} & & \textbf{32.49} & \textbf{49.82} \\
   FedEMA (autoscaler, $\tau = 0.7$) & ResNet-18 & 11M & \textbf{73.44} & \textbf{79.49} & & \textbf{33.04} & \textbf{50.48} \\
   % FedEMA (autoscaler, $\tau = 0.8$) & ResNet-18 & 11M & \textbf{73.50} & \textbf{79.80} & & \textbf{32.91} & \textbf{50.31} \\
   \hline
   Standalone training & ResNet-50 & 23M & 63.65 & 74.30 & & 23.18 & 41.43 
   \\
   % FedAvg (Super.) & ResNet-50 & 23M & 17.72 & 21.69 & & 14.47 & 13.98 
   % \\ 
   % FedSimCLR & ResNet-50 & 23M & 26.03 & 33.83 & & 14.02 & 20.01
   % \\
   FedSimCLR & ResNet-50 & 23M & 63.00 & 73.56 & & 19.30 & 41.13 \\
   FedMoCoV1 & ResNet-50 & 23M & 61.85 & 75.53 & & 22.12 & 46.43 \\
   FedMoCoV2 & ResNet-50 & 23M & 64.25 & 73.96 & & 25.79 & 42.52 \\
   FedSimSiam & ResNet-50 & 23M & 61.46 & 15.25 & & 16.03 & 29.76 \\
   FedBYOL & ResNet-50 & 23M & 69.99 & 76.69 & & 26.57 & 45.46 \\
   FedCA \citep{zhang2020fedca} & ResNet-50 & 23M & 28.50 & 36.28 & & 16.48 & 22.46 \\
   FedU \citep{zhuang2021fedu} & ResNet-50 & 23M & 69.76 & 80.25 & & 28.42 & 48.42 \\
   FedEMA ($\lambda = 1$) & ResNet-50 & 23M & \textbf{74.64} & \textbf{81.48} & & \textbf{31.42} & \textbf{49.92} \\ 
   FedEMA (autoscaler, $\tau = 0.7$) & ResNet-50 & 23M & \textbf{72.52} & \textbf{80.68} & & \textbf{29.68} & \textbf{50.75} \\
   % FedEMA (autoscaler, $\tau = 0.8$) & ResNet50 & 23M & \textbf{73.35} & \textbf{80.24} & & \textbf{29.38} & \textbf{49.4} \\

   \hline 
   BYOL (Centralized) & ResNet-18 & 11M & 87.67 & 87.89 & & 40.96 & 56.60 \\
   BYOL (Centralized) & ResNet-50 & 23M & 89.07 & 89.66 & & 41.49 & 60.23 \\
   \hline
   \end{tabular}
   \end{center}
\end{table*}

\subsection{Network Architecture}

\textbf{Predictor} \, The network architecture of the predictor is a two-layer multilayer perceptron (MLP). The two-layer MLP starts from a fully connected layer with 4096 neurons. Followed by one-dimension batch normalization and a ReLU activation function, it ends with another fully connected layer with 2048 neurons.

\textbf{Encoder} \,  We use ResNet-18 \cite{he2016resnet} as the default network architecture of the encoder in the majority of experiments. Besides, we also provide results of ResNet-50 in Table \ref{tab:linear-eval-full} and \ref{tab:semi-sup-full}. Our ResNet architecture differs from the implementation in PyTorch \citep{paszke2017pytorch} in three aspects: 1) We use kernel size $3 \times 3$ for the first convolution layer instead of $7\times7$; 2) We use an average pooling layer with kernel size $4\times4$ before the last linear layer instead of adaptive average pooling layer; 3) We replace the last linear layer with a two-layer MLP. The network architecture of the MLP is the same as the predictor. 

% Encoder: ResNet, replace the last classifer layer with a two-layer MLP:
% 1. linear(512, 4096)
% 2. BN1d(4096)
% 3. ReLU
% 4. linear(4096, 2048)

% Predictor, a two-layer MLP:
% 1. linear(2048, 4096)
% 2. BN1d(4096)
% 3. ReLU
% 4. linear(4096, 2048)

\begin{table}[t]
   \begin{center}
   \begin{tabular}{lccccc}
      \hline
      \multicolumn{1}{l}{\multirow{2}{*}{Method}} &
      \multicolumn{5}{c}{\# of classes per client (\%)}
      \\ 
      \cline{2-6}
      & 2 & 4 & 6 & 8 & 10 (iid)
      \\
      % \# of classes per client & 2 & 4 & 6 & 8 & 10 (iid)
      \Xhline{3\arrayrulewidth}
      FedBYOL & 57.51 & 56.96 & 55.14 & 54.96 & 54.24 \\
      FedSimSiam & 48.94 & 51.08 & 49.05 & 48.09 & 49.92 \\
      FedBYOL, update-both & 49.53 & 54.17 & 51.50 & 52.70 & 53.41 \\
      \hline
   \end{tabular}
   \caption{Top-1 accuracy comparison on various non-IID levels --- the number of classes per client on the CIFAR-100 dataset. Update-both means updating both $W_k$ and $W_k^t$ with $W_g$.}
   \label{tab:update-both-cifar100}
   \end{center}
\end{table}

\begin{table*}[t]
   \begin{center}
   \begin{tabular}{lccccc}
   \hline
   \multicolumn{1}{l}{\multirow{2}{*}{Method}} &
   \multicolumn{2}{c}{CIFAR-10 (\%)} &
   \multicolumn{1}{c}{} &
   \multicolumn{2}{c}{CIFAR-100 (\%)}
   \\ 
   \cline{2-3} \cline{5-6}
   \multicolumn{1}{c}{} & IID & Non-IID & & IID & Non-IID
   \\
   \Xhline{3\arrayrulewidth}
   FedBYOL w/o EMA & 54.11 & 50.20 & & 23.82 & 25.83 \\  
   FedBYOL w/o EMA and stop-grad & 21.21 & 11.97 & & 3.74 & 2.79 \\
   FedBYOL w/o EMA and stop-grad, update-both & 82.29 & 68.75 & & 48.74 & 41.91 \\
   FedBYOL & \textbf{84.29} & \textbf{79.44} & & \textbf{54.24} & \textbf{57.51} \\ 
   \hline
   \end{tabular}
   \caption{Comparison of FedBYOL without exponential moving average (EMA) and stop-gradient (sg) on the CIFAR datasets. FedBYOL w/o EMA and sg can hardly learn, but updating both $W_k$ and $W_k^t$ with $W_g$ (update-both) enables it to achieve comparable results.}
   \label{tab:fedbyol-study}
   \end{center}
\end{table*}    

\subsection{Training and Evaluation Details}

We implement FedSSL in Python using EasyFL \citep{zhuang2022easyfl}, an easy-to-use federated learning platform based on PyTorch \citep{paszke2017pytorch}. The following are the details of training and evaluation.

\textbf{Training} \, We use Stochastic Gradient Descent (SGD) as the optimizer in training. We use $\eta = 0.032$ as the initial learning rate and decay the learning with a cosine annealing \citep{sgdr-cosine-lr}, which is also used in SimSiam. By default, we train $R = 100$ rounds with local epochs $E = 5$ and batch size $B = 128$ using $K = 5$ clients. We simulate training of $K$ clients on $K$ NVIDIA V100 GPUs and employ the PyTorch \cite{paszke2017pytorch} communication backend (NCCL) for communications between clients and the server. If not specified, we use $\lambda = 1$ by default or autoscaler with $\tau = 0.7$ for FedEMA. As for experiments of FedU, we follow the hyper-parameters described in paper \citep{zhuang2021fedu}.

\textbf{Cross-silo FL vs Cross-device FL} \, This paper primarily focuses on cross-silo FL where clients are stateful with high availability. Clients can cache local models and carry these local states from round to round. Extensive experiments demonstrate that FedEMA achieves the best performance under this setting. On the other hand, cross-device FL assumes there are millions of stateless clients that might participate in training just once. Due to the constraints of experimental settings, the majority of studies conduct experiments with at most hundreds of clients \citep{Wang2020fedma,jeong2021fedsemi}. FedEMA can work under such experimental settings by caching the states of clients in the server. When the number of clients scales to millions, FedEMA degrades to FedBYOL that updates both encoders --- without keeping any local states. 

% selection rate In such setting, F
% might take would  and   As such, clients can cache local states (local models)

\textbf{Evalaution} \, We assess the quality of learned representations using linear evaluation \citep{kolesnikov2019revisiting, grill2020byol} and semi-supervised learning \citep{zhai2019s4l, chen2020simclr} protocols. We first obtain a trained encoder (or learned representations) using full training set for linear evaluation and 99\% or 90\% of the training set for semi-supervised learning (excluding the 1\% or 10\% for fine-tuning). Then, we conduct evaluations based on the trained encoder. For linear evaluation, we train a new fully connected layer on top of the frozen trained encoder (fixed parameters) for 200 epochs, using batch size 512 and Adam optimizer with learning rate 3e-3. For semi-supervised learning, we add a new two-layer MLP on top of the trained encoder and fine-tune the whole model using 1\% or 10\% of data for 100 epochs, using batch size 128 and Adam optimizer with learning rate 1e-3. In both evaluation protocols, we remove the two-layer MLP of the encoder by replacing it with an identity function. 

% Besides, we discover a trick in linear evaluation that further improves the performance. In linear evaluation, we train a linear classifier on top of frozen features. Instead of removing the last layer (two-layer MLP) from the encoder, we replace it with an Identity mapping. Superisingly, it leads to around 1-2\% accuracy improvement in evaluation. As a result, we raise the performance of the compared methods \citep{zhuang2021fedu} compared with their original implementations. 

\begingroup
\setlength{\tabcolsep}{0.4em}
\begin{figure}[t!]
   \centering
   \begin{tabular}[c]{cc}
     \subfloat{\label{}%
         \includegraphics[width=0.4\linewidth]{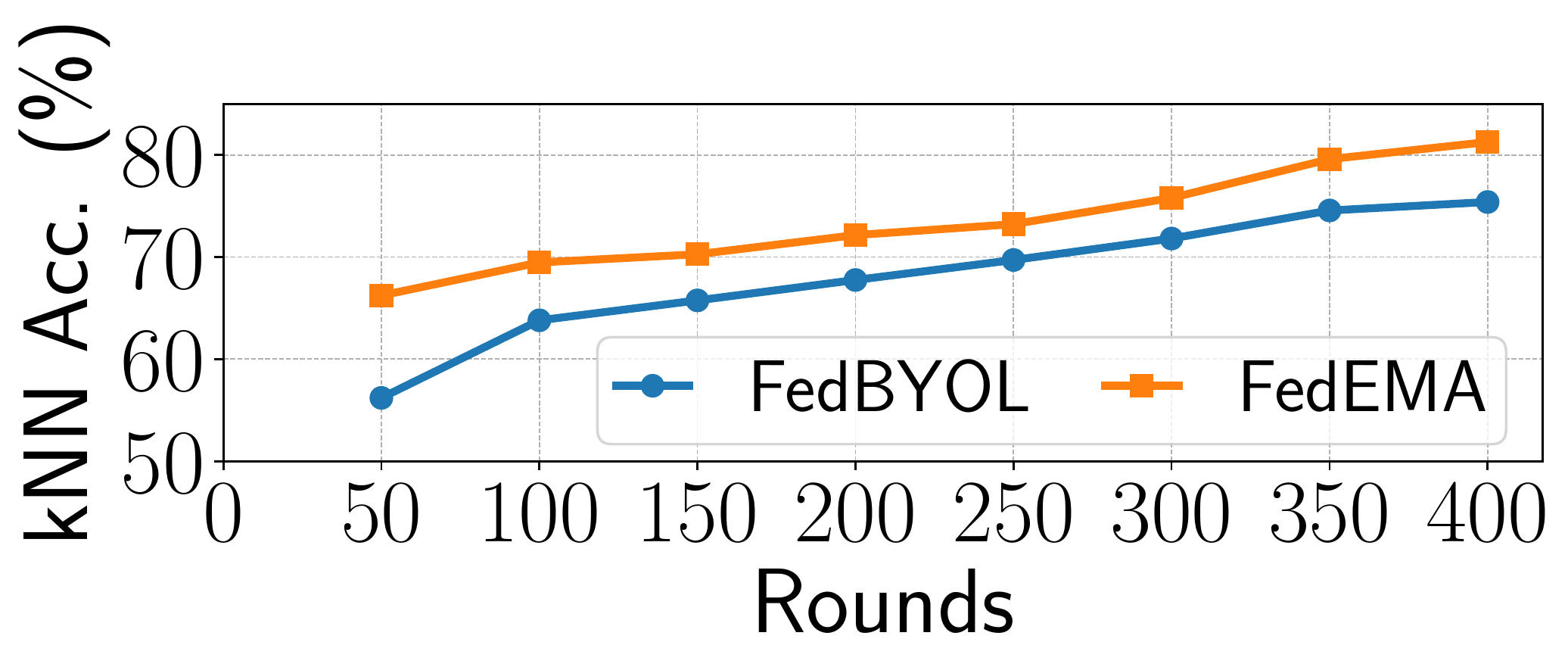}
     }
     \hfill
     \hspace{1cm}
     \subfloat{\label{}%
      \begin{tabular}{lcccc}
         \hline
         \multicolumn{1}{l}{\multirow{2}{*}{Method}} &
         \multicolumn{4}{c}{Rounds $R$ (\%)}
         \\ 
         \cline{2-5}
         & 100 & 200 & 300 & 400
         \\
         \Xhline{3\arrayrulewidth}
         FedBYOL & 79.08 & 82.23 & 83.77 & 86.09 \\  
         FedEMA (ours) & \textbf{80.78} & \textbf{82.41} & \textbf{84.08} & \textbf{86.51} \\
         \hline
      \end{tabular}
    }
   \end{tabular}
   \caption{Comparison of FedBYOL and FedEMA on various total training rounds $R$ on the non-IID setting of the CIFAR-10 dataset. FedEMA consistently outperforms FedBYOL.}
   \label{fig:rounds}
 \end{figure}
 \endgroup

\section{Additional Experimental Results and Analysis}
\label{apx:experimental-results}

In this section, we provide more experimental results of algorithm comparisons and further analyze FedEMA in different data amounts, training rounds $R$, and batch sizes $B$. 

\subsection{More Experimental Results}

Table \ref{tab:linear-eval-full} presents top-1 accuracy comparison under linear evaluation of a wide range of methods on CIFAR datasets using both ResNet-18 and ResNet-50. It supplements the algorithm comparisons in Section \ref{sec:algorithm-comparison-fedssl} and \ref{sec:algorithm-comparison-fedema}. Interestingly, FedMoCoV1 achieves good performances on IID settings of CIFAR-100 dataset. Since decentralized data are mostly non-IID, we focus more on the non-IID setting. FedEMA outperforms all the other methods in non-IID settings of CIFAR datasets. We use $\lambda = 0.8$ when using ResNet-18 and $\lambda = 1$ when using ResNet-50.

Table \ref{tab:scalability} shows results of scaling to larger numbers of clients $K$ with subsampling clients in each training round. We run two sets of experiments: 1) randomly selecting 5 out of 20 clients in each round with local epoch $E = 5$ and total rounds $R = 400$; 2) randomly selecting 8 out of 80 clients in each round with local epoch $E = 2$ and total rounds $R = 800$. We run FedEMA with autoscaler. FedEMA consistently outperforms FedBYOL with both encoders updated. We conduct these experiments using ResNet-18.

Table \ref{tab:semi-sup-full} supplements the semi-supervised learning results on Table \ref{tab:semi-sup}, providing additional results using ResNet-50 as the network architecture for the encoder. FedEMA consistently outperforms all the other methods. 

Besides, Table \ref{tab:update-both-cifar100} and \ref{tab:fedbyol-study} compare FedSimSiam, FedBYOL, and variances of FedBYOL to further demonstrate the insights from empirical studies. They supplement results in Table \ref{tab:update-both} and Figure \ref{fig:no_ema_no_sg}.

% We supplement the empirical studies with results on CIFAR-100 datasets in Table \ref{tab:update-both-cifar100} and \ref{tab:fedbyol-study}. 

% \subsection{Empirical Study}

\subsection{Further Analysis}

% \begin{minipage}{0.45\textwidth}
%    \centering
%    \includegraphics[width=1\linewidth]{charts/autoscaler.pdf}
%    \captionof{figure}{Ablation study on $\tau$ for autoscaler. Performances of $\tau \in [0.5, 1)$ are generally better the other values. We run experiments on non-IID settings using ResNet-18.}
%    \label{fig:autoscaler}
% \end{minipage}
% \hfill
% \begin{minipage}{0.45\textwidth}
%    \centering
%    \includegraphics[width=1\linewidth]{charts/ema_encoder_predictor.pdf}
%    \captionof{figure}{Top-1 accuracy of using different combinations of constant $\mu_o$ on the online encoder and constant $\mu_p$ on the predictor. Although good choices of such combination yield better performance than FedBYOL, our proposed FedEMA outperforms all these constant values of $\mu$.}
%    \label{fig:ema-encoder-predictor}
% \end{minipage}

\begin{figure}[t!]
   \centering
   \begin{subfigure}[t]{0.45\textwidth}
       \includegraphics[width=\textwidth]{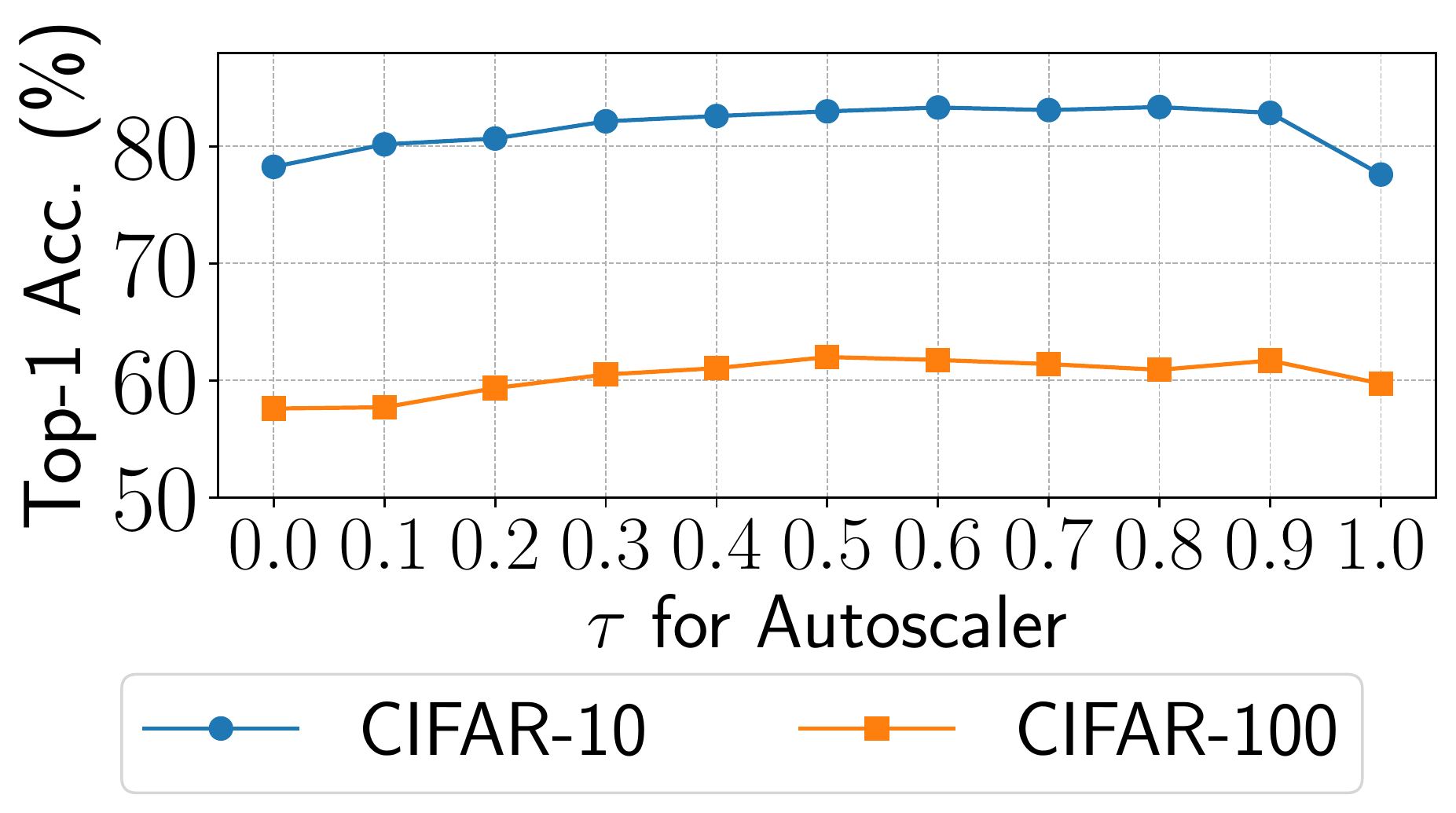}
       \caption{$\tau$ for autoscaler}
       \label{fig:autoscaler}
   \end{subfigure}
   \hfill
   \begin{subfigure}[t]{0.5\textwidth}
      \includegraphics[width=\textwidth]{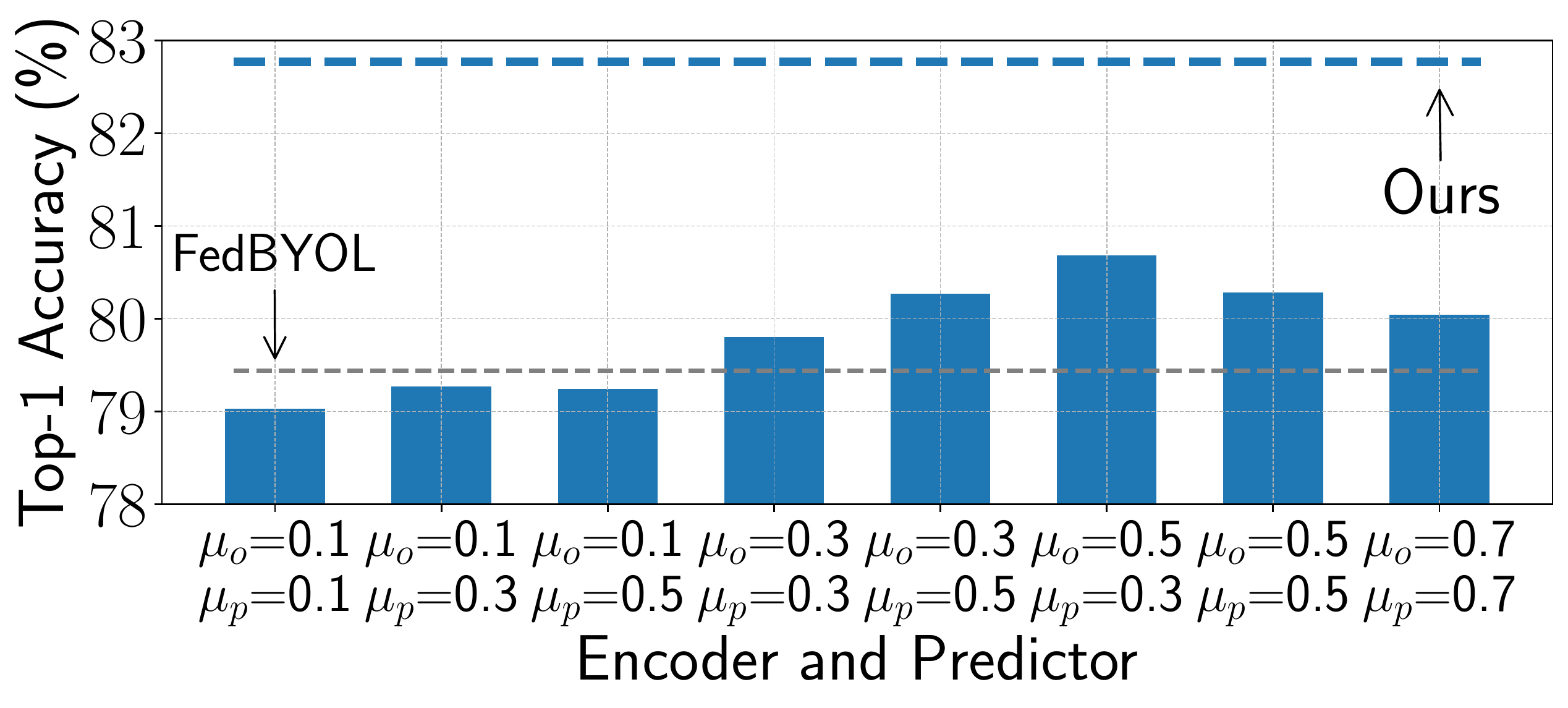}
      \caption{Combinations of $\mu_o$ and $\mu_p$}
      \label{fig:ema-encoder-predictor}
  \end{subfigure}
  \caption{Ablation study on $\tau$ for autoscaler and combinations of constant $\mu$: (a) analyzes the impact of $\tau$ on performances; 2) presents top-1 accuracy of using different combinations of constant $\mu_o$ on the online encoder and constant $\mu_p$ on the predictor.}
  \label{fig:ablation-apx}
\end{figure}

% \begin{figure}[t]
%    \begin{center}
%       \includegraphics[width=0.5\textwidth]{charts/autoscaler.pdf}
%       \caption{Ablation study on $\tau$ for autoscaler. Performances of $\tau \in [0.5, 1)$ are generally better the other values. We run experiments on non-IID settings using ResNet-18.}
%       \label{fig:autoscaler}
%    \end{center}
% \end{figure}

% \begin{figure}[t]
%    \begin{center}
%       \includegraphics[width=0.5\textwidth]{charts/ema_encoder_predictor.pdf}
%       \caption{Top-1 accuracy of using different combinations of constant $\mu_o$ on the online encoder and constant $\mu_p$ on the predictor. Although good choices of such combination yield better performance than FedBYOL, our proposed FedEMA outperforms all these constant values of $\mu$.}
%       \label{fig:ema-encoder-predictor}
%    \end{center}
% \end{figure}

\begin{table}[t]
   \begin{center}
   \begin{tabular}{lcccccc}
      \hline
      \multicolumn{1}{l}{\multirow{2}{*}{Method}} &
      \multicolumn{5}{c}{Batch Sizes $B$ (\%)}
      \\ 
      \cline{2-7}
      & 16 & 32 & 64 & 128 & 256 & 512
      \\
      \Xhline{3\arrayrulewidth}
      FedBYOL & 68.74 & 72.90 & 78.58 & 79.44 & 79.80 & 77.74 \\
      % FedEMA (ours) (Manual tuned lambda) & \textbf{79.55} & \textbf{81.31} & \textbf{82.77} & \textbf{82.70} & \textbf{79.93} \\
      % FedEMA (autoscaler, $\lambda=0.8$) & 73.21 & \textbf{77.83} & \textbf{82.13} & \textbf{83.26} & \textbf{82.51} & \textbf{80.79} \\
      FedEMA (ours) & \textbf{74.03} & \textbf{79.06} & \textbf{82.18} & \textbf{83.34} & \textbf{82.19} & \textbf{80.51} \\

      \hline
   \end{tabular}
   \caption{Top-1 accuracy comparison on various batch sizes $B$ on the non-IID setting of CIFAR-10 dataset. The batch size should not be either too small or too large. Besides, FedEMA outperforms FedBYOL.}
   \label{tab:batch-size}
   \end{center}
\end{table}

\begin{table*}[t!]
   \caption{Comparison of needed communication rounds to reach target accuracy using different local epochs $E$ on the non-IID setting of the CIFAR-10 dataset. $E = 1$ is unable to reach 80\% in 100 rounds. A larger $E$ can reduce communication costs by increasing the computation cost.}
   \label{fig:comp_vs_comm}
   \begin{center}
      \begin{tabular}{cccccccccc}
         \hline
         \multicolumn{1}{c}{\multirow{2}{*}{Target accuracy}} &
         \multicolumn{4}{c}{Communiation (rounds)} & &
         \multicolumn{4}{c}{Computation (epochs)}
         \\
         \cline{2-5} \cline{7-10}
         \multicolumn{1}{l}{} & E = 1 & E = 5 & E = 10 & E = 20 & & E = 1 & E = 5 & E = 10 & E = 20 \\
         \Xhline{3\arrayrulewidth}
         70\% & 90 & 40 & 10 & 8 & & 90 & 200 & 100 & 160 \\ 
         80\% & - & 80 & 50 & 40 & & - & 400 & 500 & 800 \\ 
         \hline
     \end{tabular}  
   \end{center}
\end{table*}

\begin{table}[t]
\begin{tabular}{l|egha|h|g}
   \hline
   \multicolumn{1}{l|}{\# of clients} &
   \multicolumn{4}{c|}{K = 5} &
   % \multicolumn{1}{c}{} &
   \multicolumn{1}{c|}{K = 10} & 
   \multicolumn{1}{c}{K = 20}
   \\ 
   \hline 
   % \cline{2-3} \cline{5-6}
   \multicolumn{1}{l|}{Data amount} & 10\% & 25\% & 50\% & 100\% & 50\% & 25\%
   \\
   % \hline 
   \Xhline{3\arrayrulewidth}
   FedBYOL & 43.27 & 65.14 & 76.11 & 78.25 & 75.10 & 63.95 \\ 
   \hline 
   FedEMA (ours)& 44.33 & 67.46 & 79.49 & 82.54 & 79.20 & 66.61 \\ 
   \hline
\end{tabular}
\caption{Top-1 accuracy comparison of various data amounts in clients and different numbers of clients. Increasing the number of clients does not improve performance, whereas increasing the data amount of clients results in better performance.}
 \label{tab:data-amount}
\end{table}

\textbf{$\tau$ for Autoscaler} \, We analyze the impact of $\tau$ on performances in Figure \ref{fig:autoscaler}. Generally, using autoscaler with $\tau \in (0, 1)$ is better than FedBYOL ($\tau = 0$). The performance of $\tau = 1$ yields worse results because only local knowledge are used in model update (the global knowledge is neglected) as discussed in Section \ref{sec:fedema}. Besides, performances of $\tau \in [0.5, 1)$ are generally better other values, which verifies our intuition discussed in Section \ref{sec:fedema}. These results also show that we can achieve even higher performance on the CIFAR-100 dataset on Table \ref{tab:linear-eval} by tunning $\tau$. We run experiments on non-IID settings using ResNet-18.

\textbf{Constant $\mu$} \, To further illustrate the effectiveness of our dynamic EMA, we provide results of using different combinations of constant $\mu_o$ on the online encoder and constant $\mu_p$ on the predictor in Figure \ref{fig:ema-encoder-predictor}. The results of $\mu_o = 0.9$ and $\mu_p = 0.9$ is only 54.52\%, which is far lower than the others. Among these combinations, $\mu_o = 0.5$ and $\mu_p = 0.3$ achieve the best performance. It suggests that better performances may be achieved if we can construct different dynamic $\mu$ for the encoder and the predictor, while we leave this interesting insight for future exploration. Although good choices of constant $\mu_o$ and $\mu_p$ achieve better performance than FedBYOL, FedEMA consistently outperforms all these methods. These results complement Figure \ref{fig:const-ema} in the main manuscript.

\textbf{Impact of Training Rounds $R$} \, Figure \ref{fig:rounds} compares FedBYOL and FedEMA with increasing number of training (communication) rounds $R$. Performances of both FedBYOL and FedEMA increases as training proceeds and FedEMA consistently outperforms FedBYOL. We run these experiments with $\lambda = 0.5$ for FedEMA on the non-IID setting of CIFAR-10 dataset. 

\textbf{Impact of Batch Size $B$} \, We investigate the impact of batch size in Table \ref{tab:batch-size}. The performances of batch size $B = 128$ and $B = 256$ are similar, outperforming the other batch sizes. It indicates that the batch size should not be either too small or too large. Besides, FedEMA outperforms FedBYOL in all batch sizes. We run the experiments with autoscaler ($\tau = 0.7$) on the non-IID setting of the CIFAR-10 dataset.
% $\lambda = 0.4$ for $B = 32$ and $\lambda = 0.6$ for $B = 64$

\textbf{Communication vs Computation Cost} \, Table \ref{fig:comp_vs_comm} shows the needed communication rounds and computation epochs to reach a target accuracy using different local epochs $E$ with FedEMA. Increasing $E$ reduces communication cost as the needed rounds decrease, but it generally requires a higher computation cost. For example, compared with $E = 5$ that needs 80 rounds to reach 80\% with 400 epochs of computation, $E = 20$ only uses 40 rounds but needs 800 epochs computation cost. These results indicate the trade-off between communication cost and computation cost.

\textbf{Data Amount} \, Table \ref{tab:data-amount} shows that increasing the data amount improves the performance significantly. By default, we split the CIFAR-10 dataset into 5 clients, each client contains 10,000 training images, denoting as 100\% data amount. As a result, $p$\% data amount means that each client contains $10,000 * p$\% images. For example, 25\% data amount means that each client contains 2,500 images. With lesser data points in each client, we can construct more clients to conduct training as the total data amount is fixed. Table \ref{tab:data-amount} shows that when the data amount is same in clients, increasing the number of clients in each training round do not improve performance. However, increasing the data amount in each client increases the performance significantly. These results indicate that it is important for clients to have sufficient data to participate in training in FedSSL.
% The overall performance is comparable as long as each client contains around 5,000 data points.

% \begin{table}[t]
%    \begin{center}
%    \begin{tabular}{lcccc}
%       \hline
%       \multicolumn{1}{l}{\multirow{2}{*}{Method}} &
%       \multicolumn{4}{c}{Rounds $R$ (\%)}
%       \\ 
%       \cline{2-5}
%       & 100 & 200 & 300 & 400
%       \\
%       \Xhline{3\arrayrulewidth}
%       FedBYOL & 79.08 & 82.23 & 83.77 & 86.09 \\  
%       FedEMA & \textbf{80.78} & \textbf{82.41} & \textbf{84.08} & \textbf{86.51} \\
%       \hline
%    \end{tabular}
%    \caption{Top-1 accuracy comparison on various total training rounds $R$ on the non-IID setting of CIFAR-10 dataset. }
%    \label{tab:rounds}
%    \end{center}
% \end{table}

% \begin{table}
%    \begin{tabular}{l|c|c|c|c}
%       % \hline
%       Method & 100 r & 200 r & 300 r & 400 r \\
%       % \hline 
%       \Xhline{3\arrayrulewidth}
%       FedBYOL & 79.08 & 82.74 & 84.96 & 86.09 \\  
%       FedEMA (0.5) & 80.78 & 82.41 & 84.08 & 86.51 \\
%       FedEMA (ours) & 81.28 & 83.51 & 85.07 & 84.83 \\ 
%       % \hline
%    \end{tabular}
%    \caption{Rounds. $\lambda = 0.5$}
%     \label{fig:rounds}
% \end{table}

\end{document}